\documentclass{article}

\usepackage{microtype}
\usepackage{graphicx}
\usepackage{subcaption}
\usepackage{booktabs}
\usepackage{dsfont}
\usepackage{makecell}
\usepackage{float}
\usepackage[dvipsnames]{xcolor}
\usepackage{bm}

\newcommand{\stats}[2]{\makecell{\small #1 \\ \footnotesize #2}}

\usepackage{hyperref}

\usepackage[preprint]{icml2026}

\usepackage{amsmath}
\usepackage{amssymb}
\usepackage{mathtools}
\usepackage{amsthm}
\usepackage{soul}
\usepackage{enumitem}
\usepackage{hyperref}

\usepackage[capitalize,noabbrev]{cleveref}

\theoremstyle{plain}

\theoremstyle{definition}

\theoremstyle{remark}

\usepackage[textsize=tiny]{todonotes}

\icmltitlerunning{Temporally Interpretable Differentiable Decision Trees}

\begin{document}

\twocolumn[
  \icmltitle{Temporally Interpretable Differentiable Decision Trees}



  \icmlsetsymbol{equal}{*}

  \begin{icmlauthorlist}
    \icmlauthor{Eisuke Hirota}{yyy,sch}
    \icmlauthor{Aarav Sane}{yyy}
    \icmlauthor{Rohan Paleja}{yyy}
  \end{icmlauthorlist}

  \icmlaffiliation{yyy}{Department of Computer Science, Purdue University}
  \icmlaffiliation{sch}{Thomas Lord Department of Computer Science, University of Southern California}

  \icmlcorrespondingauthor{Eisuke Hirota}{\texttt{eh$\_$352@usc.edu}}

  \icmlkeywords{Machine Learning, ICML}

  \vskip 0.3in
]



\printAffiliationsAndNotice{}  % no special notice (required even if empty)

\begin{abstract}
  Interpretability offers a solution to safe autonomy by providing transparency into an agent's underlying decision-making model.
  Within sequential-decision making tasks, differentiable decision trees (DDTs) are one approach to such interpretability, maintaining automatic-differentiable policies while providing humans with a discrete tree-based visualization. Nonetheless, current implementations of DDTs are not well-suited for sequential-decision making domains, as there exists an inherent mismatch between a tree's single-timestep behavior and a human's multi-timestep planning.
  Our work thus introduces \textit{time} as a new dimension of interpretability, coined as temporal interpretability, \textcolor{black}{and demonstrates how temporal abstractions via action chunking improve it.}
  We achieve this by first introducing two novel policy gradient algorithms that incorporate action chunking. Additionally, to maintain parameter-efficient trees, we develop an information-theoretic tree restructuring algorithm that modifies the tree during training.
  Across four simulation environments, we find that warm-starting action chunked DDTs from a distilled action chunked policy is the most effective way to obtain temporally interpretable trees: they match neural network policies in three of the four domains while using up to 80$\%$ fewer parameters.
  Our code is available at \url{https://github.com/ei5uke/temp-interp}.
\end{abstract}

\section{Introduction}
% introduce why we use interpretable AI
Societal applications such as healthcare, autonomous transportation, and criminal justice must promote interpretability and cannot afford delegating final decision-making to black-box models. By handing over the wheel, we risk causing life-changing instances like medical black-boxes returning with severely incorrect diagnoses \cite{zech2018variable}, autonomous vehicles driving with extraordinary risk and user burden \cite{9346000}, and criminal defendants receiving unjustifiably long sentences determined by opaque models \cite{stevenson2018algorithmic}. Human input must be taken into account before real-world deployment of these models, e.g., debugging model behavior via the opinions of engineers, researchers, or domain-specific professionals, or determining the legality of integrating such models within current litigation. Interpretable AI provides a solution to such, promoting interpretable models and thus enabling a human to parse through the decision-making flow; accordingly, interpretable AI creates opportunities for easy debugging and legal analysis.
To accomplish this, interpretable AI often leverages tree-based methods, relying on its logic-based design and sparse parameter space to enable a human to easily comprehend the agent's behavior \cite{quinlan1987simplifying}.

% introduce the usage of DDTs in particular
Particularly for sequential-decision making domains, recent work in differentiable decision trees (DDTs) has improved their usage in interpretable reinforcement learning (RL) \cite{silva2019optimization, paleja2022learning, paleja2024designs}. DDTs maintain the sparsity and interpretability of traditional decision trees but enable automatic-differentiability for feasible RL. Such models are often trained directly with approximate gradients \cite{paleja2022learning}, achieving performance comparable to that of NNs.
While DDTs have enabled improved performance of trees in sequential-decision making domains, this does not directly guarantee that the model is \emph{easily} interpretable. Firstly, the logic-based tree structure within DDTs is only comprehendible if the tree is sparse: a massive flow diagram requires extensive and potentially confusing review of the entire decision pipeline \cite{hu2019optimal}. Secondly, the single-timestep action space does not provide a descriptive explanation of the agent's plans. Humans leverage logic to determine a sequence of actions or a plan, thus implying that current DDTs do not provide outputs that are easily interpretable at a plan-level, i.e., \emph{temporally} interpretable.

% introduce the method we want to use to address said problem
We thus study \emph{time} as a new dimension of interpretability and how models can be temporally interpretable (see \autoref{fig:overview}(a)). Toward temporal interpretability, we leverage action chunked RL \cite{li2025reinforcement} as a key building block. Advancements in action chunking have promoted its usage in behavior cloning, offline RL, and most recently, online RL.
Thus, we leverage action chunking to extend traditional policy gradient methods and apply it to DDTs to achieve superior interpretability in sequential decision-making paradigms.
Our key insight is the following:

% the TLDR
\begin{quote}
    \textit{action chunked tree policies can return their future actions, depicting short-term plans. This knowledge can increase user confidence in an agent's behavior or improve insight for an engineer / researcher to verify robustness and debug performance issues.}
\end{quote}

% summary and main points
We thus develop DDTs that have high temporal interpretability and aim to alleviate the increase in parameter count. To accomplish this, we develop policy gradient algorithms with action chunking to train temporally interpretable DDTs and dynamically restructure the tree during training to create compact, parameter-efficient trees.

Our main contributions are as follows:
\begin{itemize}[leftmargin=*, nosep]
    \item We introduce two novel policy gradient methods with action chunking, developing agents that can accurately showcase their short-term plan to the user at each timestep, thus increasing behavior transparency.
    \item We introduce Information-Theoretic Tree Restructuring (ITTR), an algorithm that deepens / prunes the tree based on policy complexity during training. Our algorithm remedies the increase in parameter count due to action chunking, reducing tree size by up to $50\%$.
    \item We demonstrate the effectiveness of our approach across four domains through extensive simulation experiments, robustness verification analysis, and deepening / pruning analysis. Warm-starting our action chunked DDTs from a distilled action chunked policy yields our best temporally interpretable trees, performing competitively against state-of-the-art baselines while using up to 80$\%$ fewer parameters.
\end{itemize}

\section{Related Works}
In this section, we discuss relevant work in the literature on interpretable AI and action chunking.

\subsection{Interpretable AI}
% introduce interpretable AI stuff
Interpretable AI comprises techniques with global model transparency \cite{9346000, atakishiyev2024explainable, zeng2019end, doshi2017towards, du2019techniques}. A human should be able to \textit{interpret} the contents of the model; they must comprehend the parameters, the logic flow, and subsequently, the agent's behavior. As such, interpretability is often achieved through leveraging models with clear logical structures like decision trees \cite{olaru2003complete, breiman2017classification} with mind to their size \cite{silva2021encoding} or rule lists \cite{weiss1995rule, angelino2018learning}, given that humans can confidently and easily understand their behavior. On the other hand, high-dimensional parameterizations like neural networks (NNs) have significantly more computations and nonlinearity, making it difficult for a human to trace the logic flow, and also leverage embedding spaces that have no clear meaning to a human overseer. Aiming to combat these obstacles, there exists some work that directly condition neural networks to be more interpretable, hoping to combine the strong generalization power of neural networks with the easy comprehension found in interpretable AI methods \cite{krakovna2016increasing, wu2021optimizing, wu2018beyond, ross2018improving}. These approaches generally achieve this by enforcing structural constraints to simplify the model's logic or applying gradient regularization so that the model learns intelligible features.

% introduce xAI as a general umbrella field
Interpretable AI is related to but distinct from explainable AI (xAI). Generally, while interpretable AI focuses on transparent models, xAI prioritizes models that can \textit{explain} its thought process \cite{linardatos2020explainable, arrieta2020explainable, adadi2018peeking}. This may include direct methods like prompt engineering LLMs to explain their thought process \cite{cambria2024xai} to more indirect methods like saliency maps to highlight significant pixels in images \cite{borys2023explainable}. Crucially, while these xAI methods provide a glimpse into the model's thought process, they do not fully resolve the mystery within black-box parameterizations, leading to a false sense of security regarding such explanations. Hence, research in interpretable AI is essential and aims to develop near-white box methods with similar performance to deep learning approaches. As such, to avoid the pitfalls of xAI, we focus on generating interpretable models that can provide a clear logical trace for their actions without relying on post-hoc explanations.

% differentiate temporal interpretability from prev works
Our work differs from prior work by introducing \textit{time} as a new dimension of interpretability. While model transparency is important in pre-deployment to determine a model's viability, online temporal comprehension is useful to understand model behavior during deployment, especially in sequential-decision-making domains, e.g., robotics control. With temporal interpretability, engineers / researchers can debug an agent's decision-making behavior before deployment, and users can take better-informed online decisions, like intervening mid-trajectory.

\subsection{Action Chunking}
% introduce AC work in robotics
Action chunking is in essence a temporal extension of policies \cite{li2025reinforcement}. Traditionally, a policy $\pi(a_t| s_t)$ outputs one action given the current state. Action chunked policies extend the action space by a time horizon $H$, thus the updated policy follows: $\pi(a_{t:t+H}| s)$. Since the policy outputs $H$ actions, there exist different paradigms regarding how often to query the policy. The simplest approach would be to query the policy every $H$ timesteps, solely relying on each outputted action chunk (i.e., open-loop). An approach commonly leveraged for deploying vision-language-action models (VLAs) on robots is Temporal Ensemble, where the policy is queried every timestep and the current action is chosen using the weighted average of the $k$ lastly outputted action chunks \cite{black2025realtime}, albeit such forced temporal consistency may be disadvantageous in highly stochastic environments. Action chunking is effective in capturing non-Markovian behaviors found within offline datasets, hence its common use in training VLAs \cite{zhao2023learning}. Lastly, action chunking has been extended to RL, in both offline \cite{li2025reinforcement, li2025decoupled} and online paradigms \cite{hahnaction}.

% Overall, action chunking has seen the most success in capturing non-Markovian behaviors found within offline datasets, hence why it is commonly used to train VLAs \cite{zhao2023learning}. Lastly, action chunking has been extended to RL, in both offline \cite{li2025reinforcement, li2025decoupled} and online paradigms \cite{hahnaction}. \textcolor{red}{Last few sentences could use some smoothing. You have a ``Overall", followed by a ``Lastly"}

% introduce AC work in cogsci
% From a cognitive science perspective, action chunking has been uncovered to aid in policy compression in humans \cite{lai2021policy, lai2025action}. Particularly, studies have found that humans readily use action chunking to lower the cognitive load of solving repetitive tasks, thus causing their overall policy to require less memory usage.

Our work thus aims to extend policy gradient algorithms through leveraging action chunking. Particularly, we study how action chunking enables agents to depict their short-term plans to users and whether techniques like temporal ensemble provide performance and interpretability benefits.

\begin{figure*}[t]
    \centering
    \includegraphics[width=\linewidth]{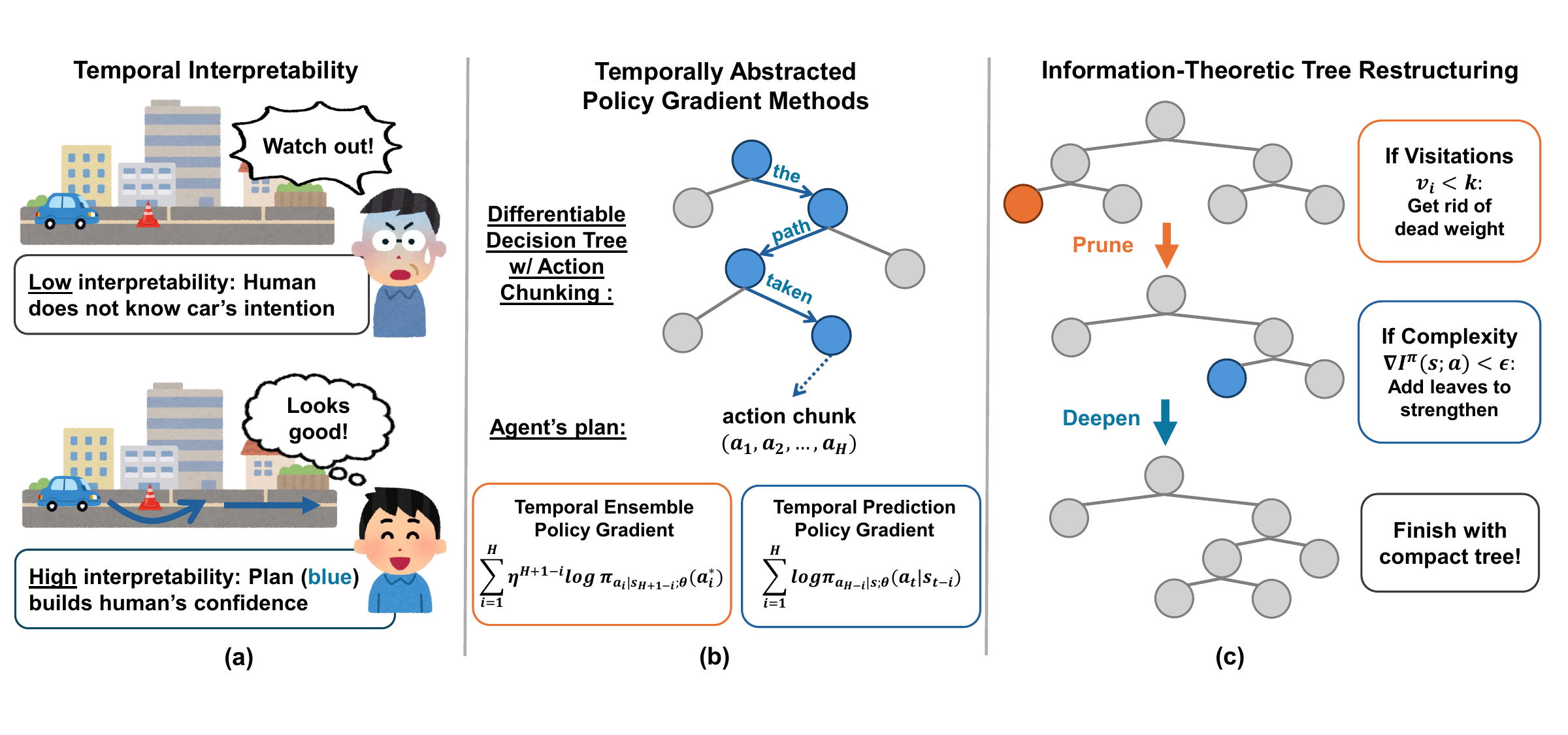}
    \caption{Overview figure. (a) Temporal Interpretability. We provide an example depicting how humans increase in confidence when provided with an agent's short-term plan and temporal decision-making (an autonomous vehicle's plan is visualized with blue arrows). (b) Temporally Abstracted Policy Gradients. We introduce two novel policy gradient methods that enable action chunked DDTs to be temporally interpretable. (c) Information-Theoretic Tree Restructuring. Introducing action chunking to trees increases parameter count, hence we dynamically deepen or prune leaves to ensure the model only contains sub-branches we need most.}
    \label{fig:overview}
\end{figure*}

\section{Preliminaries}
\subsection{Problem Formulation}
We define our problem as a Markov Decision Process (MDP), a 6-tuple $\langle S,A,P,R,\gamma,p_0\rangle$ where $S$ is the state space, $A$ is the action space, $P:S\times A\rightarrow S$ is the transition probability, $R:S\times A\rightarrow \mathbb{R}$ is the reward function, $\gamma\in[0,1]$ is the discount factor, and $p_0:S\rightarrow[0,1]$ is the initial state distribution. We use RL to learn the optimal policy $\pi^*(\mathrm{a}_t|\mathrm{s}_t)=\text{argmax}_\pi \mathbb{E}_{\tau\sim\pi}[\sum^T_{t=0}\gamma^t R(\mathrm{s}_t,\mathrm{a}_t)]$ that maximizes the expected return from the MDP, where $\tau=\langle \mathrm{s}_0,\mathrm{a}_0,...,\mathrm{s}_T,\mathrm{a}_T\rangle$ is the agent's episodic trajectory.

\subsection{Differentiable Decision Trees}
\textcolor{black}{DDTs are similar to traditional decision trees but maintain differentiability, enabling learning via backpropagation} \cite{817409, silva2019optimization, paleja2022learning}. This is achieved through leveraging fuzzy logic in place of previously binary decision points. Particularly, DDTs perform a sigmoid function during a decision point $i$:
\begin{equation}
\label{eq:ddt}
    y_i = \frac{1}{1+\exp(-\alpha (w^T_i x - b_i)},
\end{equation}
where $w_i$ are the weights of the decision node, $x$ is the feature vector of the current state, $b_i$ is the splitting criterion, $\alpha$ is the sigmoid's steepness hyperparameter, and $y_i$ is the probability of the node $i$ outputting TRUE.

This architecture itself is not interpretable, however, due to probabilistic splitting. Hence, DDTs are generally converted to DTs using crispification. This process works by only allowing the parameter with the largest magnitude absolute weight in each weight vector to determine the splitting, changing \autoref{eq:ddt} to: $y_i = \mathds{1}(w^k_i x^k - b_i > 0)$, where $k$ is the index of the parameter with the most absolute weight.

For our experiments, we use the ICCT variant due to its increased performance and interpretability \cite{paleja2022learning}. Generally, trees lose quality during the crispification process as comparisons within each node lose important features. The ICCT variant solves this problem through directly optimizing over the crisp tree using the straight-through trick \cite{bengio2013estimating}, creating interpretable trees that can be trained directly and parsed through.

\subsection{Policy Gradient}
Policy gradient methods work by directly adjusting the parameters of a policy instead of querying argmaxed actions from a Q-value function to act as a proxy of a policy \cite{sutton1998reinforcement}. Given a policy $\pi$ parameterized by $\theta$, policy gradient works by minimizing the following:
\vspace{-1pt}
\begin{equation}
\label{eq:pg}
    J(\theta) = \mathbb{E}_\theta \left[ \log\pi(\mathrm{a}_t|\mathrm{s}_t)\hat A_t \right],
\end{equation}
where $\hat A_t$ is the approximated advantage function, typically following something akin to:
\begin{equation}
\label{eq:advantage}
    \hat A_t = r_t + \gamma V_{\phi}(\mathrm{s}_{t+1}) - V_{\phi}(\mathrm{s}_t).
\end{equation}
Here, $r_t$ is the reward signal at time $t$, $\gamma$ is the discount factor,  and $V_\phi(\mathrm{s})$ is the value function parameterized by $\phi$, often learned via minimizing its distance to the true value $V^*_t$ calculated via the recursive Bellman optimality equation.

Adding action chunking to policy gradient algorithms was introduced recently and is straightforward \cite{hahnaction}. The goal is to temporally extend the policy to output action chunks, $\mathrm{a}_{t:t+H}=\{\mathrm{a}_t, \mathrm{a}_{t+1},\ldots,\mathrm{a}_{t+H-1}\}$, while the value function is left alone since it does not take any actions. The policy gradient equations are then updated (in blue):
\begin{gather}
\label{eq:pg-ac}
    J(\theta) = \mathbb{E}_\theta \left[ \log\pi(\mathrm{a}_{\textcolor{blue}{t:t+H}}|\mathrm{s}_t)\hat A_{\textcolor{blue}{t+H}} \right], \\
    \hat{A}_{\textcolor{blue}{t+H}} = \textcolor{blue}{\sum_{t'=t}^{t+H-1} \gamma^{t'-t} r_{t'}} + \gamma^{\textcolor{blue}{H}} V_\phi(s_{\textcolor{blue}{t+H}}) - V_\phi(s_t),
\end{gather}
where $H$ is the action chunking horizon. For clarity, we provide the reader with a derivation of this policy gradient in \autoref{sec:deriv}.

% the problem w/ naively using basic PG w/ AC
Unfortunately, such a policy provides little benefit in terms of temporal interpretability because this policy is trained to be queried every $H$ timesteps. Hence, the user only sees an action chunk every $H$ timesteps rather than continuously, leading to situations where information for timesteps $\geq t+H+1$ are missing during timesteps $t:t+H$. While such policy can still be deployed, querying at every timestep instead of $H$ timesteps, this will likely decrease agent performance and lead to unstable trajectories as the agent is only trained to account for transitions $P(\mathrm{s}_{kH}, \mathrm{a}_{kH})$.

% \subsection{Information Theory}
% % introduce info theory for later use of policy complexity
% The study of deep NNs has often leveraged information theory to gain insights into learning. Particularly, past literature has modeled human action and cognition as an RL problem focused on achieving high performance with minimum data \cite{lai2021policy}. This objective of minimizing policy complexity is modeled as the mutual information $I$ between the chosen state-action pairs by a given policy $\pi$:

% \begin{align}
% \begin{split}
%     I^\pi (\mathrm{s};\mathrm{a}) &= H(\mathrm{s})-H(\mathrm{s}|\mathrm{a}) \\
%     &= \sum_s p(\mathrm{s}) \sum_\mathrm{a} \pi(\mathrm{a}|\mathrm{s})\log \frac{\pi(\mathrm{a}|\mathrm{s})}{p(\mathrm{a})},
% \end{split}
% \end{align}

% where $H$ is Shannon's entropy. Policies with high complexity have low conditional entropy $H(\mathrm{s}|\mathrm{a})$, intuitively implying that the policy focuses on mapping unique state-action pairs. On the other hand, policies with low complexity may behave in a less injective manner; for example, a completely random policy is the least complex policy possible.

% We leverage policy complexity in our dynamic deepening / pruning algorithm. Particularly, inspired by literature highlighting increases in policy complexity in human behavior to imply increases in cognitive demand, we hypothesize that increases in policy complexity of trees to imply an increase in task difficulty and to thus deepen and reduce leaf entropy.

\section{Method}
We introduce temporal interpretability to our DDTs through training them with novel policy gradient methods with action chunking and dynamically deepening / pruning them online. From a high-level, the idea is to enable DDTs to output their short-horizon plan to the user while compacting trees through deepening / pruning. We first introduce the policy gradient methods with action chunking (\autoref{fig:overview}(b)).

\subsection{Temporal Ensemble Policy Gradient}

\begin{algorithm}[t]
\setlength{\leftskip}{2cm}
  \setlength{\rightskip}{2cm}
   \caption{Temporal Ensemble Policy Gradient}
   \label{alg:pg-te}
\begin{algorithmic}
   \STATE {\bfseries Input:} $\theta, \phi, \mathcal{D} \leftarrow \emptyset, \eta$
   \FOR{each iteration}
   \STATE $\rho\leftarrow$ emptied queue with max length $H$
   \FOR{each environment step}
   \STATE $\mathrm{a}_{t:t+H}\sim\pi_\theta(\mathrm{a}_{t:t+H}|\mathrm{s}_t)$
    \IF{$|\rho| = H$}
    \STATE $\rho.\text{pop}(0)$
    \ENDIF
    \STATE $\rho\leftarrow\rho\cup \mathrm{a}_{t:t+H}$
   \STATE Query $\rho$ to get $\mathrm{a}^\star$ \hfill (\autoref{eq:temp-ensemble-query})
   \STATE Perform EMA with $\eta$ to compute $\mathrm{a}_t$ \hfill (\autoref{eq:temp-ensemble-main})
   \STATE $\ell_\theta(\mathrm{a}^\star) = \sum_{i=1}^{H} w_i \log \pi_{\mathrm{a}_i|s_{t-H+i};\theta}(\mathrm{a}^\star_i)$
   \STATE $\mathcal{D}\gets\mathcal{D}\cup{(s_{t-H+1:t}, \mathrm{a}^\star, r_t, V_t, \ell_\theta(\mathrm{a}^\star))}$
   \ENDFOR
   \FOR{each gradient step}
    \STATE $\mathrm{s}_{t-H+1:t}, \mathrm{a}^\star, V_t, \ell_\theta(\mathrm{a}^\star), R_t\leftarrow \mathcal{D}$.pop()
    \STATE Compute $\hat A_t$ from $V_t$ and $R_t$
    \STATE Update value parameters $\phi$ with MSE
    \STATE Update policy parameters $\theta$ \hfill(\autoref{eq:pg_temp})
   \ENDFOR
   \ENDFOR
  \STATE {\bfseries Output:} $\theta, \phi$
\end{algorithmic}
\end{algorithm}
% mention the problem w/ simply using PG w/ AC: the user only sees the action sequence every H timesteps, NOT every timestep
While the previously updated objective function enables the development of a policy with action chunking, it does not necessarily improve interpretability. Given the policy is only queried after every $H$ timesteps, the user will only see a projected plan every $H$ timesteps. If that projected plan provided little information or the user missed it, then the action chunk did not give a meaningful interpretation of the agent's behavior to the user, thus lacking the essential interpretability safety-critical autonomous systems require.

% propose a solution to the mentioned problem: temporal ensemble.
To give the user a plan, querying the agent at every timestep enables observing an action chunk at every timestep. A viable strategy is to thus develop such a policy and leverage temporal ensemble to aggregate the current action. Herein, the user will always be provided the agent's short-term plan and have increased confidence of whether to intervene or not. We coin this approach as Temporal Ensemble Policy Gradient. The implementation is detailed in Algorithm \ref{alg:pg-te} and explained below.

% introduce notation to help w/ explanation
For training such agents with temporal ensemble, we must define how such a policy works. We define our chunking policy $\pi_\theta(\mathrm{a}_{t:t+H}|\mathrm{s}_t)$ which outputs an action chunk with horizon $H$ upon inputting the current state. This chunking policy is queried at every timestep, and we use a history of the last $H$ chunks to perform temporal ensemble and determine the current action. Given a history of chunks $\rho=\{\mathrm{a}^\dagger_{t-H+1}, \ldots, \mathrm{a}^\dagger_{t}\}$, where $\mathrm{a}^\dagger_{k}$ is the action chunk sampled at timestep $k$, temporal ensemble is performed by firstly selecting the actions that each chunk assigns to timestep $t$:
\vspace*{-2pt}
\begin{equation}
\label{eq:temp-ensemble-query}
    \mathrm{a}^\star = \{\mathrm{a}^\dagger_{t-H+1,H}, \mathrm{a}^\dagger_{t-H+2,H-1},\ldots,\mathrm{a}^\dagger_{t,1}\},
\end{equation}
where $\mathrm{a}^\dagger_{t,H}$ represents the $H^{th}$ element in chunk $\mathrm{a}^\dagger_t$. We then proceed to take the exponential moving average (EMA) over the actions in the history where $\mathrm{a}^\star_i$ corresponds to each element in $\mathrm{a}^\star$ and $\eta\in(0,1]$ is a coefficient term for EMA:
\vspace*{-2pt}
\begin{equation}
    \label{eq:temp-ensemble-main}
    \mathrm{a}_t = \frac{\sum_{i=1}^{H} \eta^{H+1-i}\mathrm{a}^\star_i}{\sum_{j=1}^{H} \eta^{H+1-j}},
\end{equation}
The policy gradient objective function is updated to:
\begin{equation}
\label{eq:pg_temp}
        \mathbb{E}_\theta\left[\underbrace{\textcolor{blue}{\sum_{i=1}^{H} w_i} \log \pi_{\textcolor{blue}{\mathrm{a}_i \mid s_{t-H+i}};\theta}(\textcolor{blue}{\mathrm{a}^\star_i)}}_{\text{temporal ensemble}}\hat{A}t\right], \quad w_i = \frac{\eta^{H+1-i}}{\sum^H_{j=1} \eta^{H+1-j}} 
\end{equation}
Since the aggregated action combines actions sampled from $H$ different chunks, we aggregate their log probabilities with the same normalized EMA weights, evaluating each $\mathrm{a}^\star_i$ under the chunk distribution of the state $s_{t-H+i}$ that produced it.  Note that this weighted sum is not the exact log-likelihood of $\mathrm{a}_t$, so \autoref{eq:pg_temp} is a surrogate objective, i.e., a biased estimator of the policy gradient for $\eta<1$. For $\eta=1$ it reduces, up to scale, to an unbiased estimator given unbiased advantage estimates, since each $\mathrm{a}^\star_i$ only affects the environment from timestep $t$ onward. Subsequently, Temporal Ensemble Policy Gradient incorporates temporal ensemble over action chunks into policy gradient, introducing temporal interpretability and giving users insight into the future actions.

\subsection{Temporal Prediction Policy Gradient}
\begin{algorithm}[t]
   \caption{Temporal Prediction Policy Gradient}
   \label{alg:pg-tp}
\begin{algorithmic}
   \STATE {\bfseries Input:} $\theta, \phi, \mathcal{D} \leftarrow \emptyset, \lambda$
   \FOR{each iteration}
   \FOR{each environment step}
   \STATE $\mathrm{a}_{t:t+H}, \log\pi(\mathrm{a}_{t:t+H}|\mathrm{s}_t)\sim\pi_{\theta}(\mathrm{a}_{t:t+H}|\mathrm{s}_t)$
   \STATE $\mathcal{D}\leftarrow\mathcal{D}\cup\{(\mathrm{s}_t, \mathrm{a}_{t:t+H}, r_t,V_t,\log\pi_{\theta}(\mathrm{a}_{t:t+H}|\mathrm{s}_t)\}$
   \ENDFOR
   \FOR{each gradient step}
       \STATE $\mathrm{s}_{t-H+1}, \mathrm{a}_t, V_t, \log\pi_\theta(\mathrm{a}_{t:t+H}|\mathrm{s}_t), R_t\leftarrow \mathcal{D}$.pop()
   \STATE Compute $\hat A_t$ from $V_t$ and $R_t$
  \STATE Update value parameters $\phi$ with MSE
  \STATE Update policy parameters $\theta$ \hfill(\autoref{eq:pg-pred})
   \ENDFOR
   \ENDFOR
  \STATE {\bfseries Output:} $\theta, \phi$
\end{algorithmic}
\end{algorithm}

% introduce the problem in temp-ensemble: bad for stochastic envs.
Temporal ensemble policy gradient provides higher temporal consistency within environments, albeit with two noticeable flaws. Firstly, enforced temporal consistency may lead to worse performance in highly stochastic environments, due to the non-Markovian nature of the objective function. Secondly, the user does not gain much temporal interpretability from the action chunk if the EMA coefficient is low. This is because any actions the agent plans to use for later will have minuscule weight when leveraging a small EMA coefficient, having very little impact and thus not providing a significant description to the user at the current timestep. Such low EMA coefficients may also cause the policy to output gibberish or noise for the future action chunks.

% deploy action chunking as a prediction problem instead of a control problem
An improved strategy is to instead condition the action chunk to predict the agent's future actions, i.e., the action chunk consists of the action at the current timestep, plus future predicted actions. This strategy, we call temporal prediction, focuses on improved temporal interpretability and adaptability to stochastic environments, in exchange for not enforcing temporal consistency like temporal ensemble. We build upon autonomous vehicle domains in which a user interface often depicts the vehicle's local plan to the user \cite{atakishiyev2024explainable, WaymoYt}.
We write this Temporal Prediction Policy Gradient in Algorithm \ref{alg:pg-tp}.

% explain the math behind the algorithm
The chunks $\{\mathrm{a}_t, \mathrm{a}_{t+1}, \ldots, \mathrm{a}_{t+H}\}$ from our chunking policy $\pi_\theta(\mathrm{a}_{t:t+H}|\mathrm{s}_t)$ are now used as follows: $\mathrm{a}_t$ is directly used for the current timestep $t$, and the subsequent actions $\mathrm{a}_{t+i}$ are only shown to the user and not applied for agentic control. Herein, the policy is trained on two objectives, one for high performance and another for showing accurately predicted future actions. The first objective is virtually equivalent to traditional policy gradient, but uses the marginal log probability of the first action $\mathrm{a}_t$ only:
\vspace*{-2pt}
\begin{equation}
    L_1(\theta) = \log\pi_{\textcolor{blue}{\mathrm{a}_1|\mathrm{s}};\theta}(\mathrm{a}_t|\mathrm{s}_t)\hat A_t.
\end{equation}This conditions the policy's first action $\mathrm{a}_t$ to successfully tackle the actual RL task. The remainder of the chunk is then conditioned to accurately predict the policy's future actions. This is achievable through conditioning each subsequent marginal log probability to match the action at timestep $t$ but with a state observed in a previous timestep:
\vspace*{-2pt}
\begin{equation}
    L_2(\theta)=\underbrace{\textcolor{blue}{\sum_{i=1}^{H-1}\log\pi_{\mathrm{a}_{i+1}|\mathrm{s};\theta}(\mathrm{a}_t|\mathrm{s}_{t-i})}}_\text{temporal prediction objective}.
\end{equation}
We only include past states $\mathrm{s}_{t-i}$ from the same episode as $\mathrm{s}_t$. Since the subsequent parts of the chunk output action $\mathrm{a}_t$ given a prior state $\mathrm{s}_{t-i}$, this is equivalent to the policy predicting a future action $\mathrm{a}_{t+i}$ given the current state $\mathrm{s}_t$. All in all, the complete objective is given:
\begin{equation}
\label{eq:pg-pred}
    L(\theta)=L_1(\theta)+\lambda L_2(\theta),
\end{equation}where $\lambda$ is a scaling coefficient. This combined loss function aims to steer policies to be reactive and comprehend what actions to perform next.
Accordingly, Temporal Prediction Policy Gradient leverages action chunks to depict predicted future actions, further improving temporal interpretability and providing users with accurate short-term plans.

\subsection{Information-Theoretic Tree Restructuring}

% introduce the parameter increase problem
The addition of action chunking has increased the policy's parameter space significantly. To remedy this, we introduce Information-Theoretic Tree Restructuring (ITTR), an algorithm that adds / cuts leaves from the DDT during training (shown in \autoref{fig:overview}(c)). Our algorithm extends the pruning design from \cite{paleja2024designs} and incorporates information theory from \cite{lai2021policy, lai2025action}. We explain the process below and refer the reader to \autoref{sec:restructure} for Algorithm \ref{alg:deep-prune}. By the best of our knowledge, we are the first to introduce an algorithm that both deepens and prunes a DDT online, during training.

% deepening
Primarily, we aim to restructure our tree such that we can maintain or gain performance with fewer parameters. This requires deepening to improve performance and pruning to cut dead weight. We first deepen a tree when the rewards are low and the parameterization is stuck. Specifically, we model this by computing whether the average gradient of the policy complexities $I^\pi(\mathrm{s};\mathrm{a})$ is less than some epsilon $\epsilon$, where $I$ is mutual information (see \autoref{alg:deep-condition}). Fulfilling such means that the tree has not learned a ``highly injective'' state-action mapping, implying that the policy uses similar actions for vastly different states. This, in conjunction with a low reward, suggests adding leaves to improve performance.
\vspace*{-3pt}
\begin{equation}
\label{alg:deep-condition}
    \frac{1}{N}\sum \nabla I^\pi(\textrm{s};\textrm{a}) < \epsilon.
\end{equation}We can then add leaves by splitting a highly-used, but lowly-varied leaf into two leaves, found by solving \autoref{alg:deep-leaf}. Here, $p(l)$ is the probability of reaching leaf $l$ and $\mu_l$ is the output of leaf $l$. Intuitively, solving this optimization problem finds the leaf responsible for causing low policy complexity. This implies that we can split this leaf into two, introducing a new comparator node to experiment with different decision splits and increasing model capacity.
\begin{equation}
\label{alg:deep-leaf}
    l^* = \underset{l}{\mathrm{argmax}} \mathbb{E}_{\mathrm{s}_t,\mathrm{a}_t,{\hat A}_t \sim D} \left[
    p(l|\mathrm{s}) * \underbrace{1/\text{Var}\left[(\mathrm{a}_t-\mu_l)\hat A_t\right]}_\text{low action uncertainty}
    \right].
\end{equation}

% note about deepening
% Note that a simpler method to choose which leaf to deepen is to simply pick the leaf with the most visitations. Unfortunately, such method fails when the leaf with the most visitations is precise, and thus provides confident outputs and should not be changed. The introduction of entropy ensures that we pick leaves that are suitable for change.

% pruning
For pruning, we simply keep track of the number of leaf visitations, $v=\{v_1, \ldots, v_l\}$, and only prune leaves with visitations less than some constant: $v_i < k$. Note that upon pruning a leaf, we also replace the leaf's parent decision node with the leaf's sibling sub-tree. Subsequently, ITTR enables our trees to efficiently expand or reduce its representations in practice, providing trees with just enough parameters to successfully perform on a task.

\section{Experiments and Results}
In this section, we describe the simulation experiments we run and the results we achieve.

\subsection{Simulation Environments and Mutual Information}
% introduce the envs
We validate our approach in \textcolor{black}{four} environments. The environments are: Leurent's \textsc{Lane Keeping} (LK) environment \cite{leurent2018environment}; Gymnasium's \textsc{Inverted Pendulum v5} (IP), \textsc{Lunar Lander v3} (LL), and \textsc{Lunar Lander v3 Hard} (LL-H) \cite{towers2024gymnasium}. All environments have default parameters except LL-H where we add multiple modifications intended to increase stochasticity and force rapid adaptations from the agent: (1) the \texttt{wind} parameter is set to 20, (2) the \texttt{turbulence} parameter is set to 2.0, and (3) we add Gaussian noise $\mathcal{N}(0,5\mathrm{e}{-3})$ to the agent's action at every timestep. The environments are considered ``solved'' with episodic returns of: 100 for LK, 1000 for IP, 200 for LH, and 100 for LL-H. Since all of these environments have continuous observation and action spaces, calculating entropy and mutual information for ITTR is intractable in nature. Instead, we estimate mutual information using InfoNCELoss (equation 11 of the appendix in \cite{oord2018representation}) and apply this value to \autoref{alg:deep-condition}. 

\subsection{Algorithms and Baselines}
% introduce the methods
For analysis, we compare \textcolor{black}{fifteen} methods:

\begin{itemize}[leftmargin=*, nosep]
    \item \textsc{MLP:} A sparse MLP w/o any temporal interpretability.
    \item \textsc{MLP-Ensemble:} \textsc{MLP} with Temporal Ensemble.
    \item \textsc{MLP-Prediction:} \textsc{MLP} with Temporal Prediction.
    \textcolor{black}{
    \item \textsc{MLP-Big:} A big MLP w/o any temporal interpretability.
    \item \textsc{MLP-Big-Ensemble:} \textsc{MLP-Big} with Temporal Ensemble.
    \item \textsc{MLP-Big-Prediction:} \textsc{MLP-Big} with Temporal Prediction.
    \item \textsc{CART:} The state-action pairs from \textsc{MLP-Big} are then distilled into a DT with linear models at its leaves using CART \citep{loh2011classification}.
    \item \textsc{CART-Ensemble:} \textsc{CART} with \textsc{MLP-Big-Ensemble} and Temporal Ensemble.
    \item \textsc{CART-Prediction:} \textsc{CART} with \textsc{MLP-Big-Prediction} and Temporal Prediction.
    \item \textsc{DDT:} A DDT without any temporal interpretability.
    \item \textsc{DDT-Ensemble:} \textsc{DDT} with Temporal Ensemble.
    \item \textsc{DDT-Prediction:} \textsc{DDT} with Temporal Prediction.
    \item \textsc{Warm:} The DT from \textsc{CART} is converted into a DDT with the same crisp splits and leaf models, and then trained with RL and ITTR as for DDT.
    \item \textsc{Warm-Ensemble:} \textsc{Warm} with Temporal Ensemble.
    \item \textsc{Warm-Prediction:} \textsc{Warm} with Temporal Prediction.
    }
\end{itemize}

We use sparse MLPs with hidden layers $[16,16]$ to match the size of trees without compromising performance. We specifically use ICCT-Complete, a variant of ICCT with linear sub-controllers leveraging all features at leaves \cite{paleja2022learning}. 
Additionally, we use Robust Policy Optimization (RPO) for our policy gradient variant due to its recorded high performance across many simulation environments \cite{rahman2022robust, huang2022cleanrl}. For the action chunk horizon length, we set $H=10$ aiming to provide more information to the user. We set $\eta=0.75$ for our Temporal Ensemble methods, chosen as a middle ground of performance and interpretability (more info in \autoref{sec:ablations}). Note that our proposed methods work for any DDT and policy gradient variant.

\begin{table*}[t]
    \scriptsize
    % \footnotesize
    \centering
    \caption{Performance comparison of all algorithms in the four simulation domains, across three runs in an evaluation environment separate from the training environment. We report two values in each cell: the top value depicts the episodic return (mean$\pm$s.e.) and the bottom value depicts the number of parameters (constant values for MLPs and mean$\pm$s.e. for DDTs). For CART, we count a feature index and a threshold per decision node and the weights and bias of each leaf's linear model.}
    % \begin{tabular}{l l l l l l l}
    \begin{tabular}{l c c c c}
        \toprule[1pt]
        & \multicolumn{4}{c}{\textbf{Environments}} \\
        \cmidrule(lr){2-5}
         \textbf{\textbf{Algorithms}} & LK & IP & LL & LL-H\\
         \midrule[1pt]
         \textsc{MLP} & \stats{$185.3\pm2.9$}{497 } & \stats{$1000.0\pm0.0$}{369} & \stats{$293.8\pm2.3$}{450} & \stats{$272.1\pm7.2$}{$450$} \\
         \midrule
         \textsc{MLP-Ensemble} & \stats{$190.2\pm2.0$}{650} & \stats{$1000.0\pm0.0$}{522} & \stats{$296.0\pm3.2$}{756} & \stats{$270\pm3.2$}{$756$}\\
         \midrule
         \textsc{MLP-Prediction} & \stats{$193.3\pm0.9$}{650} & \stats{$1000.0\pm0.0$}{522} & \stats{$294.6\pm1.6$}{756 }& \stats{$273.2\pm5.6$}{$756$} \\
         \midrule
        \textcolor{black}{\textsc{MLP-Big}} & \textcolor{black}{\stats{$191.4\pm1.6$}{$5057$}} & \textcolor{black}{\stats{$1000.0\pm0.0$}{$4545$}} & \textcolor{black}{\stats{$292.4\pm3.8$}{$4866$}} & \textcolor{black}{\stats{$269.6\pm4.0$}{$4866$}}\\
         \midrule
       \textcolor{black}{\textsc{MLP-Big-Ensemble}} & \textcolor{black}{\stats{$189.2\pm3.8$}{$5642$}} & \textcolor{black}{\stats{$1000.0\pm0.0$}{$5130$}} & \textcolor{black}{\stats{$293.3\pm270$}{$6036$}} & \textcolor{black}{\stats{$274.3\pm3.2$}{$6036$}} \\
         \midrule
       \textcolor{black}{\textsc{MLP-Big-Prediction}} & \textcolor{black}{\stats{$193.7\pm0.2$}{$5642$}} & \textcolor{black}{\stats{$1000.0\pm0.0$}{$5130$}} & \textcolor{black}{\stats{$296.1\pm1.8$}{$6036$}} & \textcolor{black}{\stats{$269.6\pm5.2$}{$6036$}}\\
         \midrule[1pt]
             \textcolor{black}{\textsc{CART}} & \textcolor{black}{\stats{$191.3\pm1.5$}{$28.0\pm0.0$}} & \textcolor{black}{\stats{$1000.0\pm0.0$}{$12.0\pm0.0$}} & \textcolor{black}{\stats{$275.3\pm3.5$}{$38.0\pm0.0$}} & \textcolor{black}{\stats{$125.1\pm8.7$}{$78.0\pm0.0$}}\\
         \midrule
               \textcolor{black}{\textsc{CART-Ensemble}} & \textcolor{black}{\stats{$167.5\pm14.2$}{$262.0\pm0.0$}} & \textcolor{black}{\stats{$787.0\pm173.9$}{$102.0\pm0.0$}} & \textcolor{black}{\stats{$149.1\pm35.3$}{$362.0\pm0.0$}} & \textcolor{black}{\stats{$129.4\pm18.6$}{$726.0\pm0.0$}}\\
         \midrule
               \textcolor{black}{\textsc{CART-Prediction}} & \textcolor{black}{\stats{$193.6\pm0.1$}{$262.0\pm0.0$}} & \textcolor{black}{\stats{$1000.0\pm0.0$}{$102.0\pm0.0$}} & \textcolor{black}{\stats{$281.1\pm1.5$}{$362.0\pm0.0$}} & \textcolor{black}{\stats{$175.9\pm8.5$}{$726.0\pm0.0$}}\\
         \midrule
        \textsc{DDT} & \stats{$193.7\pm1.2$}{$55.7\pm9.5$} & \stats{$1000.0\pm0.0$}{$20.0\pm0.0$} & \stats{$287.8\pm0.9$}{$50.0\pm0.0$} & \stats{$247.5\pm3.5$}{$75.3\pm20.7$}\\
         \midrule
         \textsc{DDT-Ensemble} & \stats{$178.7\pm11.5$}{$278.0\pm0.0$} & \stats{$1000.0\pm0.0$}{$110.0\pm0.0$} & \stats{$259.3\pm5.5$}{$374.0\pm0.0$} & \stats{$160.8\pm13.1$}{$440.0\pm53.9$}\\
         \midrule
          \textsc{DDT-Prediction} & \stats{$162.2\pm13.8$}{$328.7\pm41.4$} & \stats{$1000.0\pm0.0$}{$110.0\pm0.0$} & \stats{$297.0\pm2.5$}{$374.0\pm0.0$} & \stats{$229.2\pm9.1$}{$506.0\pm53.9$}\\
         \midrule[1pt]
               \textcolor{black}{\textsc{Warm}} & \textcolor{black}{\stats{$195.4\pm1.0$}{$44.0\pm0.0$}} & \textcolor{black}{\stats{$1000.0\pm0.0$}{$20.0\pm0.0$}} & \textcolor{black}{\stats{$298.2\pm0.8$}{$50.0\pm0.0$}} & \textcolor{black}{\stats{$258.7\pm3.3$}{$171.3\pm21.7$}}\\
         \midrule
               \textcolor{black}{\textsc{Warm-Ensemble}} & \textcolor{black}{\stats{$195.2\pm0.5$}{$278.0\pm0.0$}} & \textcolor{black}{\stats{$1000.0\pm0.0$}{$110.0\pm0.0$}} & \textcolor{black}{\stats{$284.7\pm5.7$}{$374.0\pm0.0$}} & \textcolor{black}{\stats{$258.4\pm1.9$}{$506.0\pm53.9$}}\\
         \midrule
               \textcolor{black}{\textsc{Warm-Prediction}} & \textcolor{black}{\stats{$195.5\pm0.5$}{$278.0\pm0.0$}} & \textcolor{black}{\stats{$1000.0\pm0.0$}{$110.0\pm0.0$}} & \textcolor{black}{\stats{$294.3\pm1.1$}{$374.0\pm0.0$}} & \textcolor{black}{\stats{$246.2\pm3.3$}{$573.3\pm94.3$}}\\
         \bottomrule[1pt]
    \end{tabular}
    \label{tab:main_results}
\end{table*}

% Obs / action space of each domain:
% LK: obs space (12), action space (1)
% IP: obs space (4), action space (1)
% LL: obs space (8), action space (2)
% SP: obs space (60), action space (2)

% \begin{table*}[t]
%     \centering
%     \caption{Performance comparison of all algorithms in the four simulation environments, across three runs. All results depict the mean$\pm$s.e. episodic return in a evaluation environment, separate from the training environment.}
%     \begin{tabular}{l l l l l}
%         \toprule
%         & \multicolumn{4}{c}{\textbf{Simulation Environments}} \\
%         \cmidrule(lr){2-5}
%          \textbf{Algorithms} & LK & IP & LL & LL-H\\
%          \midrule
%          \textsc{MLP} & \\
%          \textsc{MLP-Ensemble} & \\
%           \textsc{MLP-Prediction} & \\
%          \textsc{DDT} & \\
%          \textsc{DDT-Ensemble} & \\
%          \textsc{DDT-Prediction} & $120\pm10$ & $1000\pm0$ & $250\pm50$ & $150\pm50$\\
%          \bottomrule
%     \end{tabular}
%     \label{tab:placeholder}
% \end{table*}

\subsection{Evaluation Results}
% We talk about the results comparing all methods. We should see that all algorithms are competitive in nature, but the DDTs are more interpretable.
\autoref{tab:main_results} depicts the main results comparing all algorithms across three random seeds, where the DDT models have undergone ITTR. Within each cell of the table, we report the episodic reward (top) and the number of parameters (bottom). We observe that all models and algorithms generally perform competitively, achieving successes in all problems.

We first make three observations. First, action chunking for temporal interpretability does not diminish the performance of neural network policies. Both Temporal Ensemble and Temporal Prediction only add action chunks for temporal interpretability, without significantly modifying the standard policy behavior. It is of note, however, that the parameter count explodes as expected.

Second, training action chunked DDTs from scratch is difficult. \textsc{DDT-Prediction} performs on par with other methods, however, may be seed-sensitive in LK and LL-H. Additionally, \textsc{DDT-Ensemble} drops notably to $160.8\pm13.1$ in LL-H, which matches our expectations in $\S$4.2 where the enforced temporal consistency may prove to be detrimental in highly stochastic domains. Moreover, action chunking substantially increases the number of parameters: in LL-H, the $\mathrm{\sim}75$ parameters of DDT grow to $\{440–506\}$ for the action chunked DDTs, despite ITTR pruning them from four leaves. This gives clear justification for the necessity of strong tree restructuring algorithms, as without them, action chunked trees may converge with a tremendous amount of parameters. 

Third, warm-starting resolves this. \textsc{Warm-Ensemble} and \textsc{Warm-Prediction} achieve the best or statistically indistinguishable returns among all action chunked trees in every domain, with low variance across seeds. They match the neural network policies in all environments while using $\{51–79\%\}$ fewer parameters than the MLPs with the same chunking, and reach $258.4\pm1.9$ and $246.2\pm3.3$ in LL-H, where \textsc{Warm-Ensemble} even outperforms \textsc{DDT} without action chunking ($247.5\pm3.5$). The remaining gap to the neural networks in LL-H ({12–27} reward) likely stems from the linear sub-controllers of ICCT-complete; we suggest future work on non-linear sub-controllers that preserve interpretability. Nonetheless, we see notable improvements in DDT performance with the introduction of action chunking into policy gradient methods.
% Intuitively, our trees are thus not only temporally interpretable through the action chunk, but also through understanding the splitting logic.

\begin{figure*}[t!]
    \centering
    % --- First Row ---
    \begin{subfigure}[b]{0.48\textwidth}
        \centering
        % Replace 'example-image-a' with your filename
        \includegraphics[width=\textwidth]{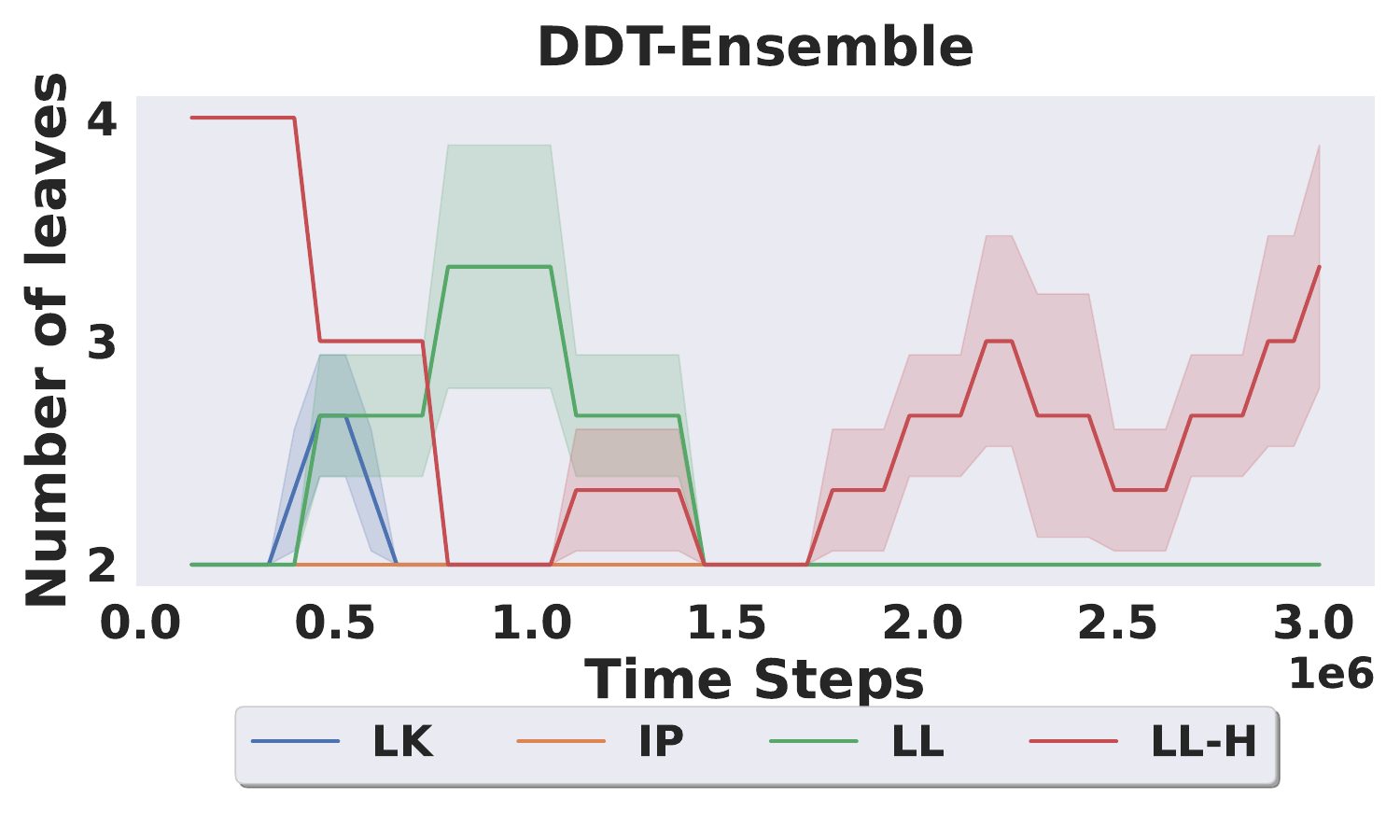}
        \caption{The trend in the number of leaves for the \textsc{DDT-Ensemble} algorithm throughout training.}
        \label{fig:result_a}
    \end{subfigure}
    \hfill % Adds flexible space between images
    \begin{subfigure}[b]{0.48\textwidth}
        \centering
        \includegraphics[width=\textwidth]{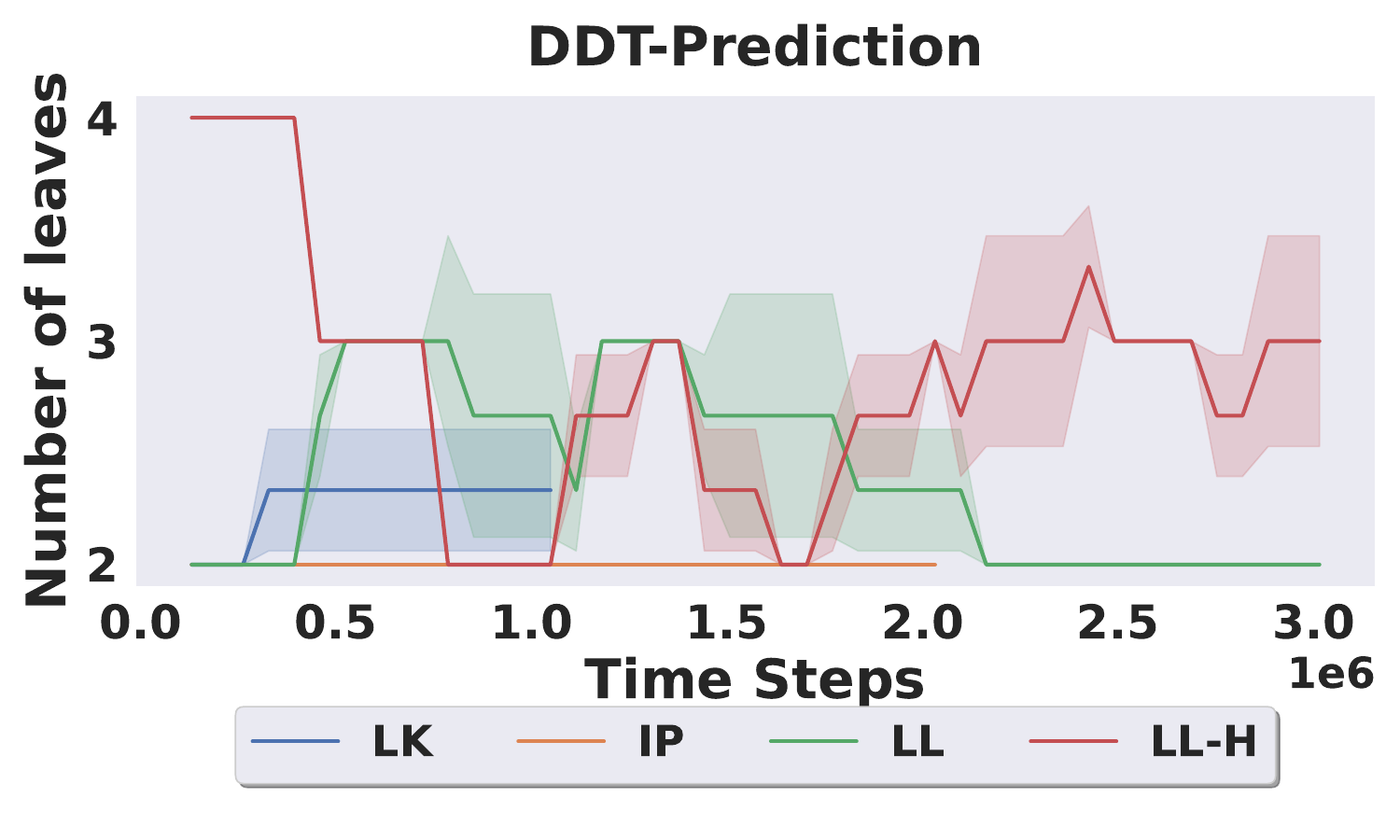}
        \caption{The trend in the number of leaves for the \textsc{DDT-Prediction} algorithm throughout training.}
        \label{fig:result_b}
    \end{subfigure}
    
    \caption{Figures depicting the trend in the number of leaves within DDTs when using our deepening / pruning algorithm for each simulation environment. Each curve showcases the mean and s.e. for a specific environment across three seeds, depicted in the dark curve and shaded regions respectively. Note that we run IP and LK for less timesteps, hence leading to the curves cutting off earlier.}
    \label{fig:restructure-results}
\end{figure*}

\subsection{Deepening / Pruning Analysis}
% introduce what this subsection is about
We also keep track of the number of leaves in \textsc{DDT-Ensemble} and \textsc{DDT-Prediction} throughout training; \autoref{fig:restructure-results} shows our results. Note that we initialize the DDTs with two leaves for LK, IP, and LL, while we use four leaves for LL-H to account for the domain's increase in difficulty.

% we see how the number of leaves have increased / decreased throughout training, properly aiming to adjust parameter count for learning.
We observe that the number of leaves fluctuates differently across the domains and runs. We observe that in IP, the trees stay constant at two leaves due to the task's easy nature. Interestingly, for LK and LL, we see a general initial increase in leaf count to only later decrease back down. This behavior appears akin to the information bottleneck \cite{shwartz2017opening}: our trees first deepen in order to have better representations and solutions, and later prunes to compress the structure and remove any insignificant information. A similar phenomena appears to exist in LL-H, albeit more varied likely due to the stochastic nature of the problem. The models may find it harder to find a compressed tree that can account for such high stochasticity.

These trends highlight the effectiveness of ITTR and how it enables effective tree learning. ITTR promotes compact trees with high interpretability in a transparent model.

\subsection{Temporal Robustness Verification}
% we input actions into a model to get future agent positions. we can show an end-user how these future positions look like, which then allows users to intervene.
With access to the future predicted actions via the proposed policy gradients, we can now perform robustness verification \cite{chen2019robustness} on the agent's short-term plans online. We can accomplish such using a closed-form model to run virtual rollouts, checking whether the action chunk predicts that the agent will exhibit unwanted termination.

\begin{figure}[h]
\centering
\begin{minipage}[h]{0.4\columnwidth}
  \centering
  % \vspace{-6pt}
  \includegraphics[width=0.8\linewidth, height=0.9\linewidth]{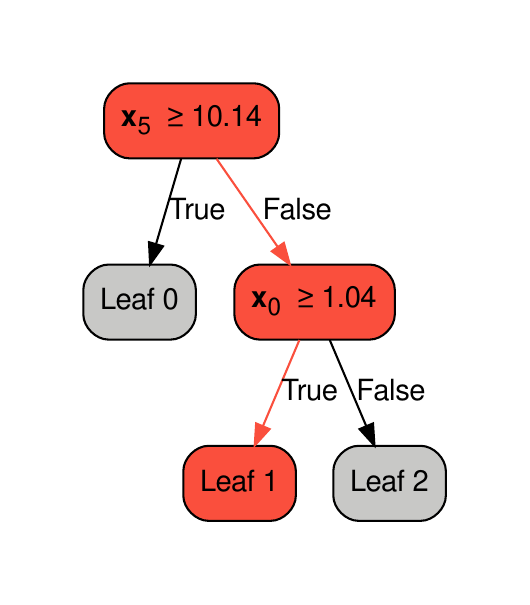}
  \captionof{figure}{A real tree's path (marked in red).}
  \label{fig:robustness_visualization}
\end{minipage}
\hfill
\begin{minipage}[h]{0.55\columnwidth}
  \centering
  % \vspace{6pt}
  \begin{tabular}{l cc}
  \toprule
    & \multicolumn{2}{c}{\textbf{Predicted}} \\
    \textbf{True} & Success & Crash \\
    \hline
    Success & 50 & 29 \\
    Crash & 2  & 15 \\
    \bottomrule
  \end{tabular}
      \captionof{table}{Confusion matrix of agent evaluation in 100 test runs (96 times completed, 4 times the agent did not predict any termination state).}
      \label{tab:confusion}
\end{minipage}
\end{figure}
We give an example in the LL-H environment in \autoref{fig:robustness_visualization} and Table \ref{tab:confusion}. Firstly, an engineer / researcher can leverage a confusion matrix (Table \ref{tab:confusion}) to assess the performance of a policy's predictions and analyze its underlying decision-making. In our example, we notice that this model has an accuracy of $67.7\%$ $(65/96)$, and is generally conservative in the remainder of guesses ($29$ out of $31$ guesses predict a crash). For sporadic domains like LL-H or autonomous vehicles where an agent can suddenly enter life-threatening situations, ensuring the creation of such conservative policies is crucial. 
The engineer may want to then simply discover ways to improve the overall accuracy, and a visualization (\autoref{fig:robustness_visualization}) depicting a path taken by the agent enables the engineer to debug the logic flow with ease. This path displays that the agent reaches leaf 1 when the agent's angular velocity is high $x_5\geq10.14$ and the agent's $x$-position is towards the right ($x_0\geq1.04$); the engineer can verify these conditions depending on a stakeholder's requirements.
Lastly, the outputted action chunks help the engineer pinpoint exactly what node the tree collapses in. This tree predicts crashes $\mathrm{\sim}{7}$ timesteps before they occur. With this in mind, the engineer can look into what happens during the final $\mathrm{\sim}7$ timesteps in each virtual rollout, verifying the viability of the agent's decision-making in these last moments. Overall, this temporal robustness verification further incentivizes real-world deployment of temporally interpretable models, as we can prevent dangerous behavior and deploy pipelines to steer agent behavior towards safe, cautionary actions.

% \subsection{Discussion}
% write further here...

\section{Conclusion}
In this work, we introduce time as a new dimension of interpretability, i.e., temporal interpretability. We present the Temporal Ensemble and Temporal Prediction policy gradient algorithms, enabling temporal abstraction within DDTs through outputting action chunks. We then ensure compact, efficient trees via an information-theoretic tree restructuring algorithm (ITTR). 
Warm-starting our action chunked DDTs from a distilled action chunked policy proves most effective: these trees match neural network policies in the LK, IP, and LL environments and come close in LL-H, while using up to 80$\%$ fewer parameters.
The inclusion of such temporal interpretability provides more incentive to realize trees and other white box methods in the real world.

\textbf{Future Work} Firstly, there exists work in state chunks, $s_{t:t+H}$. It would be interesting to extend our method such that the value function is state chunked, $V_\phi(s_{t:t+H})$, and investigate in what domains this model would be applicable towards, albeit
the non-Markovian features of Temporal Ensemble and Temporal Prediction policy gradient will likely increase training difficulty. Additionally, further research in the parameter choices of ITTR and other creative usages of information theory would be useful to create a general tree restructuring algorithm that achieves high performance across a wide range of domains. Future work can also consider the addition of various RL techniques for the proposed policy gradient methods, e.g., the usage of a scaling coefficient in Temporal Prediction, akin to the advantage mixing coefficient leveraged in \cite{fu2022deep}. Lastly, training action-chunked trees from scratch remains challenging in highly stochastic domains, and closing this gap without warm-starting is an interesting direction.

\newpage
\section{Acknowledgments}
We would like to graciously thank illustrator Takashi Mifune (pseudonym Irasutoya) for providing gratis clip art that we have used in our overview figure.

\section{Impact Statement}
This work introduces time as a new dimension of interpretability and promotes the usage of white-box models that are temporally interpretable. Such properties are particularly relevant in sequential-decision making domains where clear comprehension of a model's underlying behavior is necessary for trust, viability, and responsibility.

\bibliography{refs}

@article{shwartz2017opening,
  title={Opening the black box of deep neural networks via information},
  author={Shwartz-Ziv, Ravid and Tishby, Naftali},
  journal={arXiv preprint arXiv:1703.00810},
  year={2017}
}

@ARTICLE{817409,
  author={Suarez, A. and Lutsko, J.F.},
  journal={IEEE Transactions on Pattern Analysis and Machine Intelligence}, 
  title={Globally optimal fuzzy decision trees for classification and regression}, 
  year={1999},
  volume={21},
  number={12},
  pages={1297-1311},
  doi={10.1109/34.817409}}

@article{silva2019optimization,
  title={Optimization methods for interpretable differentiable decision trees in reinforcement learning},
  author={Silva, Andrew and Killian, Taylor and Rodriguez, Ivan Dario Jimenez and Son, Sung-Hyun and Gombolay, Matthew},
  journal={arXiv preprint arXiv:1903.09338},
  year={2019}
}

@article{paleja2022learning,
  title={Learning interpretable, high-performing policies for autonomous driving},
  author={Paleja, Rohan and Niu, Yaru and Silva, Andrew and Ritchie, Chace and Choi, Sugju and Gombolay, Matthew},
  journal={arXiv preprint arXiv:2202.02352},
  year={2022}
}

@article{li2025reinforcement,
  title={Reinforcement learning with action chunking},
  author={Li, Qiyang and Zhou, Zhiyuan and Levine, Sergey},
  journal={arXiv preprint arXiv:2507.07969},
  year={2025}
}

@article{li2025decoupled,
  title={Decoupled Q-Chunking},
  author={Li, Qiyang and Park, Seohong and Levine, Sergey},
  journal={arXiv preprint arXiv:2512.10926},
  year={2025}
}

@article{zhao2023learning,
  title={Learning fine-grained bimanual manipulation with low-cost hardware},
  author={Zhao, Tony Z and Kumar, Vikash and Levine, Sergey and Finn, Chelsea},
  journal={arXiv preprint arXiv:2304.13705},
  year={2023}
}

@article{lai2025action,
  title={Action chunking as conditional policy compression},
  author={Lai, Lucy and Huang, Ann ZX and Gershman, Samuel J},
  journal={Cognition},
  volume={264},
  pages={106201},
  year={2025},
  publisher={Elsevier}
}

@incollection{lai2021policy,
  title={Policy compression: An information bottleneck in action selection},
  author={Lai, Lucy and Gershman, Samuel J},
  booktitle={Psychology of learning and motivation},
  volume={74},
  pages={195--232},
  year={2021},
  publisher={Elsevier}
}

@ARTICLE{9346000,
  author={Chen, Jianyu and Li, Shengbo Eben and Tomizuka, Masayoshi},
  journal={IEEE Transactions on Intelligent Transportation Systems}, 
  title={Interpretable End-to-End Urban Autonomous Driving With Latent Deep Reinforcement Learning}, 
  year={2022},
  volume={23},
  number={6},
  pages={5068-5078},
  doi={10.1109/TITS.2020.3046646}}

@book{sutton1998reinforcement,
  title={Reinforcement learning: An introduction},
  author={Sutton, Richard S and Barto, Andrew G and others},
  volume={1},
  number={1},
  year={1998},
  publisher={MIT press Cambridge}
}

@inproceedings{rahman2025robust,
  title = {Robust Policy Gradient Optimization through Action Parameter Perturbation in Reinforcement Learning},
  author = {Rahman, Md Masudur and Wachs, Juan P. and Xue, Yexiang},
  booktitle = {Proceedings of the Aligning Reinforcement Learning Experimentalists and Theorists Workshop (ARLET@NeurIPS 2025) at the Thirty-Ninth Annual Conference on Neural Information Processing Systems (NeurIPS 2025)},
  year = {2025},
  note = {Peer-reviewed, non-archival workshop paper},
}

@article{hahnaction,
  title={Action Chunking Proximal Policy Optimization for Universal Dexterous Grasping},
  year = {2025},
  author={Hahn, Sanghyun and Choi, Jonghyun}
}

@article{atakishiyev2024explainable,
  title={Explainable artificial intelligence for autonomous driving: A comprehensive overview and field guide for future research directions},
  author={Atakishiyev, Shahin and Salameh, Mohammad and Yao, Hengshuai and Goebel, Randy},
  journal={IEEE Access},
  year={2024},
  publisher={IEEE}
}

@inproceedings{zeng2019end,
  title={End-to-end interpretable neural motion planner},
  author={Zeng, Wenyuan and Luo, Wenjie and Suo, Simon and Sadat, Abbas and Yang, Bin and Casas, Sergio and Urtasun, Raquel},
  booktitle={Proceedings of the IEEE/CVF conference on computer vision and pattern recognition},
  pages={8660--8669},
  year={2019}
}

@misc{WaymoYt,
	author = {Waymo},
	title = { - {Y}ou{T}ube --- youtube.com},
	howpublished = {\url{https://www.youtube.com/watch?v=nAuna_qzf6k}},
	year = {2025},
	note = {[Accessed 18-12-2025]},
}

@article{krakovna2016increasing,
  title={Increasing the interpretability of recurrent neural networks using hidden markov models},
  author={Krakovna, Viktoriya and Doshi-Velez, Finale},
  journal={arXiv preprint arXiv:1606.05320},
  year={2016}
}

@article{doshi2017towards,
  title={Towards a rigorous science of interpretable machine learning},
  author={Doshi-Velez, Finale and Kim, Been},
  journal={arXiv preprint arXiv:1702.08608},
  year={2017}
}

@article{wu2021optimizing,
  title={Optimizing for interpretability in deep neural networks with tree regularization},
  author={Wu, Mike and Parbhoo, Sonali and Hughes, Michael C and Roth, Volker and Doshi-Velez, Finale},
  journal={Journal of Artificial Intelligence Research},
  volume={72},
  pages={1--37},
  year={2021}
}

@inproceedings{wu2018beyond,
  title={Beyond sparsity: Tree regularization of deep models for interpretability},
  author={Wu, Mike and Hughes, Michael and Parbhoo, Sonali and Zazzi, Maurizio and Roth, Volker and Doshi-Velez, Finale},
  booktitle={Proceedings of the AAAI conference on artificial intelligence},
  volume={32},
  number={1},
  year={2018}
}

@inproceedings{ross2018improving,
  title={Improving the adversarial robustness and interpretability of deep neural networks by regularizing their input gradients},
  author={Ross, Andrew and Doshi-Velez, Finale},
  booktitle={Proceedings of the AAAI conference on artificial intelligence},
  volume={32},
  number={1},
  year={2018}
}

@book{breiman2017classification,
  title={Classification and regression trees},
  author={Breiman, Leo and Friedman, Jerome and Olshen, Richard A and Stone, Charles J},
  year={2017},
  publisher={Chapman and Hall/CRC}
}

@article{olaru2003complete,
  title={A complete fuzzy decision tree technique},
  author={Olaru, Cristina and Wehenkel, Louis},
  journal={Fuzzy sets and systems},
  volume={138},
  number={2},
  pages={221--254},
  year={2003},
  publisher={Elsevier}
}

@article{weiss1995rule,
  title={Rule-based machine learning methods for functional prediction},
  author={Weiss, Sholom M and Indurkhya, Nitin},
  journal={Journal of Artificial Intelligence Research},
  volume={3},
  pages={383--403},
  year={1995}
}

@article{angelino2018learning,
  title={Learning certifiably optimal rule lists for categorical data},
  author={Angelino, Elaine and Larus-Stone, Nicholas and Alabi, Daniel and Seltzer, Margo and Rudin, Cynthia},
  journal={Journal of Machine Learning Research},
  volume={18},
  number={234},
  pages={1--78},
  year={2018}
}

@article{linardatos2020explainable,
  title={Explainable ai: A review of machine learning interpretability methods},
  author={Linardatos, Pantelis and Papastefanopoulos, Vasilis and Kotsiantis, Sotiris},
  journal={Entropy},
  volume={23},
  number={1},
  pages={18},
  year={2020},
  publisher={MDPI}
}

@article{arrieta2020explainable,
  title={Explainable Artificial Intelligence (XAI): Concepts, taxonomies, opportunities and challenges toward responsible AI},
  author={Arrieta, Alejandro Barredo and D{\'\i}az-Rodr{\'\i}guez, Natalia and Del Ser, Javier and Bennetot, Adrien and Tabik, Siham and Barbado, Alberto and Garc{\'\i}a, Salvador and Gil-L{\'o}pez, Sergio and Molina, Daniel and Benjamins, Richard and others},
  journal={Information fusion},
  volume={58},
  pages={82--115},
  year={2020},
  publisher={Elsevier}
}

@article{adadi2018peeking,
  title={Peeking inside the black-box: a survey on explainable artificial intelligence (XAI)},
  author={Adadi, Amina and Berrada, Mohammed},
  journal={IEEE access},
  volume={6},
  pages={52138--52160},
  year={2018},
  publisher={IEEE}
}

@article{cambria2024xai,
  title={Xai meets llms: A survey of the relation between explainable ai and large language models},
  author={Cambria, Erik and Malandri, Lorenzo and Mercorio, Fabio and Nobani, Navid and Seveso, Andrea},
  journal={arXiv preprint arXiv:2407.15248},
  year={2024}
}

@article{borys2023explainable,
  title={Explainable AI in medical imaging: An overview for clinical practitioners--Beyond saliency-based XAI approaches},
  author={Borys, Katarzyna and Schmitt, Yasmin Alyssa and Nauta, Meike and Seifert, Christin and Kr{\"a}mer, Nicole and Friedrich, Christoph M and Nensa, Felix},
  journal={European journal of radiology},
  volume={162},
  pages={110786},
  year={2023},
  publisher={Elsevier}
}

@article{rahman2022robust,
  title={Robust policy optimization in deep reinforcement learning},
  author={Rahman, Md Masudur and Xue, Yexiang},
  journal={arXiv preprint arXiv:2212.07536},
  year={2022}
}

@article{paleja2024designs,
  title={Designs for enabling collaboration in human-machine teaming via interactive and explainable systems},
  author={Paleja, Rohan and Munje, Michael and Chang, Kimberlee and Jensen, Reed and Gombolay, Matthew},
  journal={Advances in Neural Information Processing Systems},
  volume={37},
  pages={64942--64969},
  year={2024}
}

@article{leurent2018environment,
  title={An environment for autonomous driving decision-making},
  author={Leurent, Edouard and others},
  year={2018}
}

@misc{towers2024gymnasium,
      title={Gymnasium: A Standard Interface for Reinforcement Learning Environments}, 
      author={Mark Towers and Ariel Kwiatkowski and Jordan Terry and John U. Balis and Gianluca De Cola and Tristan Deleu and Manuel Goulão and Andreas Kallinteris and Markus Krimmel and Arjun KG and Rodrigo Perez-Vicente and Andrea Pierré and Sander Schulhoff and Jun Jet Tai and Hannah Tan and Omar G. Younis},
      year={2024},
      eprint={2407.17032},
      archivePrefix={arXiv},
      primaryClass={cs.LG},
      url={https://arxiv.org/abs/2407.17032}, 
}

@article{oord2018representation,
  title={Representation learning with contrastive predictive coding},
  author={Oord, Aaron van den and Li, Yazhe and Vinyals, Oriol},
  journal={arXiv preprint arXiv:1807.03748},
  year={2018}
}

@inproceedings{fu2022deep,
  author    = {Fu, Zipeng and Cheng, Xuxin and Pathak, Deepak},
  title     = {Deep Whole-Body Control: Learning a Unified Policy for Manipulation and Locomotion},
  booktitle = {Conference on Robot Learning ({CoRL})},
  year      = {2022},
}

@article{huang2022cleanrl,
  author  = {Shengyi Huang and Rousslan Fernand Julien Dossa and Chang Ye and Jeff Braga and Dipam Chakraborty and Kinal Mehta and João G.M. Araújo},
  title   = {CleanRL: High-quality Single-file Implementations of Deep Reinforcement Learning Algorithms},
  journal = {Journal of Machine Learning Research},
  year    = {2022},
  volume  = {23},
  number  = {274},
  pages   = {1--18},
  url     = {http://jmlr.org/papers/v23/21-1342.html}
}

@article{chen2019robustness,
  title={Robustness verification of tree-based models},
  author={Chen, Hongge and Zhang, Huan and Si, Si and Li, Yang and Boning, Duane and Hsieh, Cho-Jui},
  journal={Advances in Neural Information Processing Systems},
  volume={32},
  year={2019}
}

@article{zech2018variable,
  title={Variable generalization performance of a deep learning model to detect pneumonia in chest radiographs: a cross-sectional study},
  author={Zech, John R and Badgeley, Marcus A and Liu, Manway and Costa, Anthony B and Titano, Joseph J and Oermann, Eric Karl},
  journal={PLoS medicine},
  volume={15},
  number={11},
  pages={e1002683},
  year={2018},
  publisher={Public Library of Science San Francisco, CA USA}
}

@article{du2019techniques,
  title={Techniques for interpretable machine learning},
  author={Du, Mengnan and Liu, Ninghao and Hu, Xia},
  journal={Communications of the ACM},
  volume={63},
  number={1},
  pages={68--77},
  year={2019},
  publisher={ACM New York, NY, USA}
}

@article{quinlan1987simplifying,
  title={Simplifying decision trees},
  author={Quinlan, J. Ross},
  journal={International journal of man-machine studies},
  volume={27},
  number={3},
  pages={221--234},
  year={1987},
  publisher={Elsevier}
}

@article{hu2019optimal,
  title={Optimal sparse decision trees},
  author={Hu, Xiyang and Rudin, Cynthia and Seltzer, Margo},
  journal={Advances in neural information processing systems},
  volume={32},
  year={2019}
}

@inproceedings{silva2021encoding,
  title={Encoding human domain knowledge to warm start reinforcement learning},
  author={Silva, Andrew and Gombolay, Matthew},
  booktitle={Proceedings of the AAAI conference on artificial intelligence},
  volume={35},
  number={6},
  pages={5042--5050},
  year={2021}
}

@inproceedings{
black2025realtime,
title={Real-Time Execution of Action Chunking Flow Policies},
author={Kevin Black and Manuel Y Galliker and Sergey Levine},
booktitle={The Thirty-ninth Annual Conference on Neural Information Processing Systems},
year={2025},
url={https://openreview.net/forum?id=UkR2zO5uww}
}

@article{stevenson2018algorithmic,
  title={Algorithmic risk assessments and the double-edged sword of youth},
  author={Stevenson, Megan T and Slobogin, Christopher},
  journal={Behavioral sciences \& the law},
  volume={36},
  number={5},
  pages={638--656},
  year={2018},
  publisher={Wiley Online Library}
}

@article{bengio2013estimating,
  title={Estimating or propagating gradients through stochastic neurons for conditional computation},
  author={Bengio, Yoshua and L{\'e}onard, Nicholas and Courville, Aaron},
  journal={arXiv preprint arXiv:1308.3432},
  year={2013}
}

@article{loh2011classification,
  title={Classification and regression trees},
  author={Loh, Wei-Yin},
  journal={Wiley interdisciplinary reviews: data mining and knowledge discovery},
  volume={1},
  number={1},
  pages={14--23},
  year={2011},
  publisher={Wiley Online Library}
}
\bibliographystyle{icml2026}

%%%%%%%%%%%%%%%%%%%%%%%%%%%%%%%%%%%%%%%%%%%%%%%%%%%%%%%%%%%%%%%%%%%%%%%%%%%%%%%
%%%%%%%%%%%%%%%%%%%%%%%%%%%%%%%%%%%%%%%%%%%%%%%%%%%%%%%%%%%%%%%%%%%%%%%%%%%%%%%
% APPENDIX
%%%%%%%%%%%%%%%%%%%%%%%%%%%%%%%%%%%%%%%%%%%%%%%%%%%%%%%%%%%%%%%%%%%%%%%%%%%%%%%
%%%%%%%%%%%%%%%%%%%%%%%%%%%%%%%%%%%%%%%%%%%%%%%%%%%%%%%%%%%%%%%%%%%%%%%%%%%%%%%
\appendix
\newpage
\onecolumn

\section{Outline of Appendices}
We format the appendix as follows: we provide a derivation of Policy Gradient with Action Chunking in section \ref{sec:deriv}; further explanation of Temporal Ensemble Policy Gradient in section \ref{sec:further}; depict a visualization of the trees modified by ITTR in section \ref{sec:visualization}; the ITTR algorithm design in section \ref{sec:restructure}; further explanation  of the impact of ITTR in section \ref{sec:further-restructure}; report additional ablations in section \ref{sec:ablations}; and denote miscellaneous details in section \ref{sec:misc}.

\section{Derivation of: Policy Gradient with Action Chunking}
\label{sec:deriv}
We present the reader with a high-level derivation of Policy Gradient with Action Chunking:
\begin{align*}
    \nabla_\theta J(\pi_\theta) &= \nabla_\theta \mathbb{E}\left[R(\tau)\right] \\
    &= \nabla_\theta \int_\tau P(\tau|\theta)R(\tau) & \text{(expand expectation)} \\
    &= \int_\tau \nabla_\theta P(\tau|\theta)R(\tau) & \text{(move gradient)} \\
    &= \int_\tau P(\tau|\theta)\nabla_\theta\log P(\tau|\theta)R(\tau) & \text{(log-derivative trick)} \\
    &= \displaystyle \mathop{\mathbb{E}}_{\tau\sim\pi_\theta}\left[\nabla_\theta \log P(\tau|\theta)R(\tau)\right] & \text{(return to expectation)} \\
    &= \displaystyle \mathop{\mathbb{E}}_{\tau\sim\pi_\theta}\Bigg[\Bigg(\nabla_\theta \log p_0(s_0) + \sum^{kH}_{t=0} \Big( \nabla_\theta \log P(s_{t+1}| s_t,a_{t})+ \\
    & \qquad \qquad \nabla\log\pi_\theta(a_{t}| s_t) \Big) \Bigg)R(\tau)\Bigg] & \text{(definition)} \\
    &\rightarrow \displaystyle \mathop{\mathbb{E}}_{\tau\sim\pi_\theta}\Bigg[\Bigg(\nabla_\theta \log p_0(s_0) + \sum^{kH}_{t=0} \Big( \nabla_\theta \log P(s_{\textcolor{blue}{t+H}}| s_t,a_{\textcolor{blue}{t:t+H}})+ \\
    & \qquad \qquad \nabla\log\pi_\theta(a_{\textcolor{blue}{t:t+H}}| s_t) \Big) \Bigg)R(\tau)\Bigg] & \text{(\textcolor{blue}{Action chunking})} \\
    &= \displaystyle \mathop{\mathbb{E}}_{\tau\sim\pi_\theta}\Bigg[\Bigg(0 + \sum^{kH}_{t=0} \Big(0 + \nabla\log\pi_\theta(a_{\textcolor{blue}{t:t+H}}| s_t)\Big)\Bigg) R(\tau)\Bigg] & \text{(zero gradient)} \\
    &= \displaystyle \mathop{\mathbb{E}}_{\tau\sim\pi_\theta}\left[\sum_{t=0}^{kH}\nabla_\theta \log \pi_\theta (a_{\textcolor{blue}{t:t+H}}| s_t) R(\tau)\right] & \text{(clean up)} \\
    % &=\displaystyle \mathop{\mathbb{E}}_{\tau\sim\pi_\theta}\left[\sum_{t=0}^{kH}\nabla_\theta \log \pi_\theta (a_{t:t+H}| s_t)\sum_{t'=t}^{kH}R\left(s_{t'}, a_{t'}, s_{t'+1}\right)\right] & \text{(rewards2go)} \\
    % &=\displaystyle \mathop{\mathbb{E}}_{\tau\sim\pi_\theta}\left[\sum_{t=0}^{kH}\nabla_\theta \log \pi_\theta (a_{t:t+H}| s_t)\sum_{t'=t}^{kH}\left(R\left(s_{t'}, a_{t'}, s_{t'+1}\right)-b(s_t)\right)\right] & \text{(baseline unaffects expectation)} \\
    % &=\displaystyle \mathop{\mathbb{E}}_{\tau\sim\pi_\theta}\left[\sum_{t=0}^{kH}\nabla_\theta \log \pi_\theta (a_{t:t+H}| s_t)\hat A_t\right] & \text{(replace r2go with advantage)}
\end{align*}

However, simply using total return $R(\tau)$ will likely cause training instability due to actions being scaled by reward signals in previous, unrelated timesteps. To overcome this, policy gradient methods generally scale the grad-log-prob $\nabla\log\pi$ by the rewards-to-go, $\sum_{t'=t}^{kH}R\left(s_{t'}, a_{t'}, s_{t'+1}\right)$, instead of the total return $R(\tau)$, but this also succumbs to the same time issue. Given the nature of our action chunking, instead of scaling action probabilities by the returns at each timestep, we would rather scale action probabilities by the returns at each time \emph{chunk}. This is akin to taking the advantage at timestep $t+H$:

\begin{equation*}
 \nabla_\theta J(\pi_\theta) = \displaystyle \mathop{\mathbb{E}}_{\tau\sim\pi_\theta}\left[\sum_{t=0}^{kH}\nabla_\theta \log \pi_\theta (a_{\textcolor{blue}{t:t+H}}| s_t) \hat A_{\textcolor{blue}{t+H}}\right].
\end{equation*}

Since this advantage takes place $H$ steps after the current timestep, the advantage is modified accordingly:

\begin{equation*}
    \hat{A}_{\textcolor{blue}{t+H}} = \textcolor{blue}{\sum_{t'=t}^{t+H-1} \gamma^{t'-t} r_{t'}} + \gamma^{\textcolor{blue}{H}} V_\phi(s_{\textcolor{blue}{t+H}}) - V_\phi(s_t),
\end{equation*}

Here, the environment returns $H$ reward signals between timestep $t$ and $t+H$, hence why the discount factor $\gamma$ is exponentiated as compared to equation \ref{eq:advantage}.

\section{Further explanation: Temporal Ensemble Policy Gradient}
\label{sec:further}
The main difference in \autoref{eq:pg_temp} compared to traditional policy gradient is the usage of a temporal ensemble log probability:
\begin{equation*}
\mathbb{E}_\theta\left[\underbrace{\textcolor{blue}{\sum_{i=1}^{H} w_i} \log \pi_{\textcolor{blue}{\mathrm{a}_i \mid s_{t-H+i}};\theta}(\textcolor{blue}{\mathrm{a}^\star_i)}}_{\text{temporal ensemble}}\hat{A}t\right], \quad w_i = \frac{\eta^{H+1-i}}{\sum^H_{j=1} \eta^{H+1-j}}
\end{equation*}

Standard practices in policy gradient use a multivariate Gaussian $\mathcal{N}(\mu,\Sigma)$ to depict the policy distribution, thus enabling easy sampling and computation of the log probability. We follow the same practice and use this distribution to sample action chunks $a_{t:t+H}$ instead of single actions $a_t$. Since each random variable within this distribution is independent of each other, we can still compute the log probability of a single action using the multivariate Gaussian PDF: $P(a_i)=\mathcal{N}(a_i;\mu_i,\sigma_i^2)$, where $i\in \{1,\dots,H\}$ is the action's index within the chunk. Upon collecting the log probability of each action within a temporal ensemble, we apply the linear combination of the log probabilities into the objective function, each evaluated under the chunk distribution of the state that produced it.

For ensembles early on in the episode, there will not exist enough chunks to take a complete ensemble over. For such situations, we just take the ensemble over however many chunks exist in the current history and renormalize the EMA weights over these chunks. When storing this data in the rollout buffer, we store zero vectors to ensure the buffer has no issues regarding ragged arrays. Furthermore, we employ zero masks of ensembles that include zero vectors, which also mark chunks from a previous episode.

\section{Visualization}
\label{sec:visualization}
\begin{figure*}[h]
    \centering
    % --- First Row ---
    \begin{subfigure}[b]{0.48\textwidth}
        \centering
        % Replace 'example-image-a' with your filename
        \includegraphics[width=\textwidth, height=4cm]{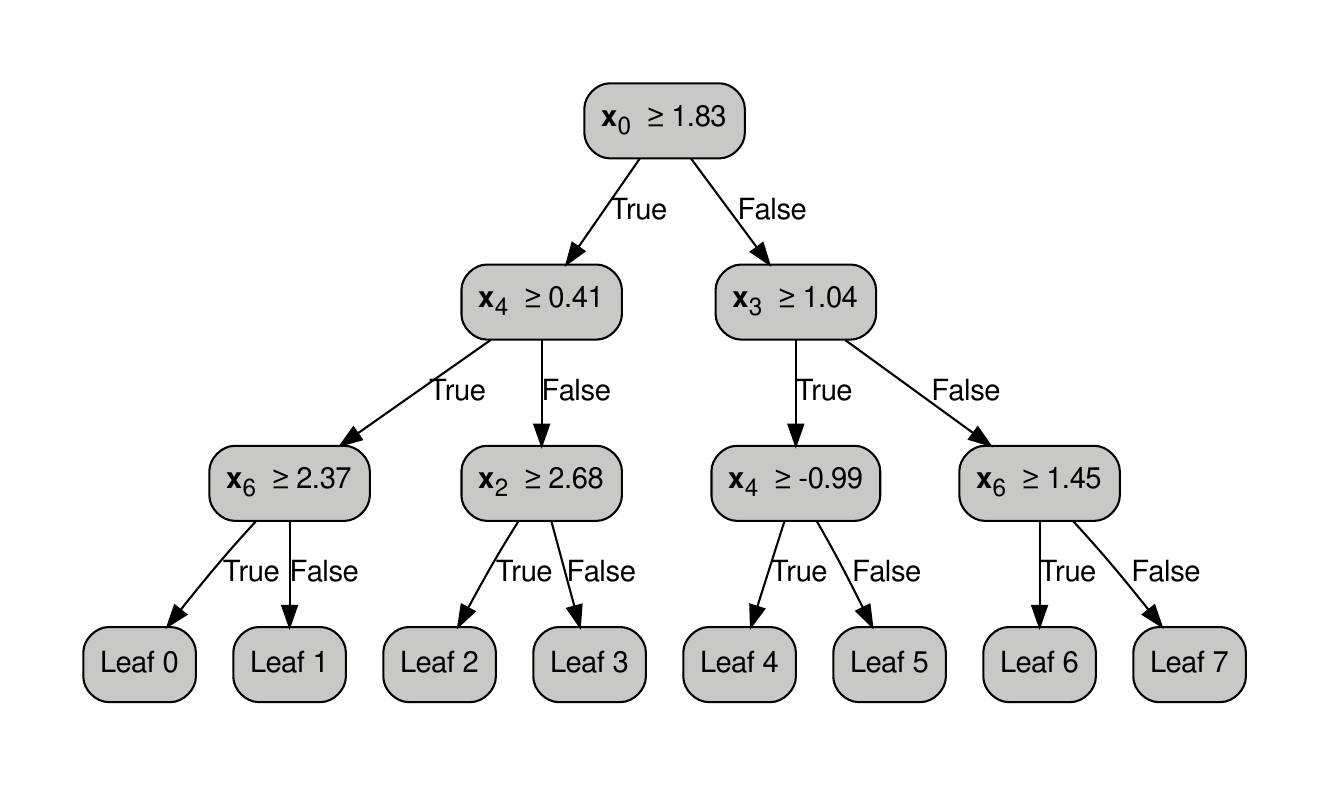}
        \caption{A tree at the beginning of training, before restructuring.}
        \label{fig:tree_start}
    \end{subfigure}
    \hfill % Adds flexible space between images
    \begin{subfigure}[b]{0.48\textwidth}
        \centering
        \includegraphics[width=0.4\textwidth, height=0.175\textheight]{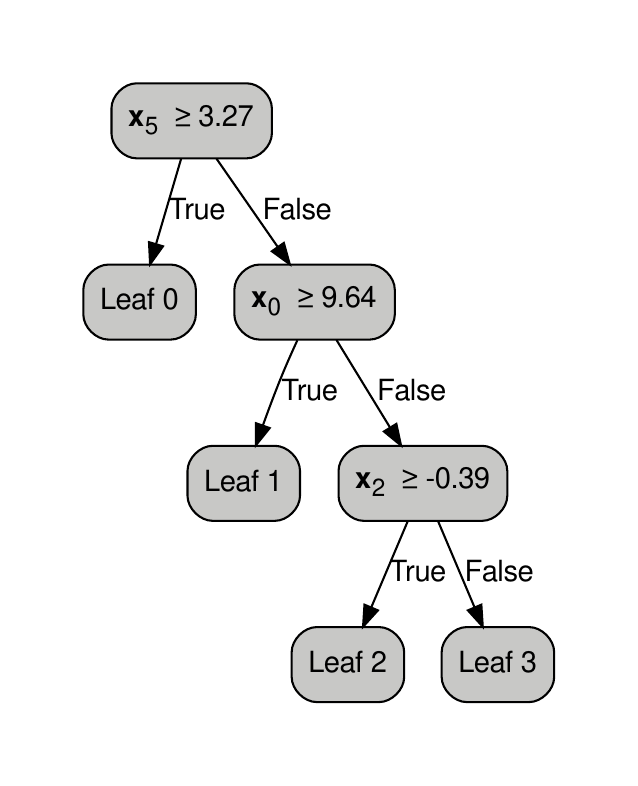}
        \caption{The saved tree with the best evaluation performance, after undergoing restructuring.}
        \label{fig:tree_end}
    \end{subfigure}
    
    \caption{Visualization of a tree at the beginning and end of a run of the LL-H domain. Each node depicts a comparison between the most significant feature in the state against a threshold.}
    \label{fig:trees}
\end{figure*}

\autoref{fig:trees} depicts a visualization of a tree in the LL-H domain, at the start and end of training where the tree underwent deepening / pruning. Particularly, these visualizations show a crisp tree, where each comparator node only chooses one feature to compare against a threshold instead of taking a linear combination of all features \cite{paleja2022learning}. Furthermore, we do not depict the contents of each leaf, as there will be significant visual clutter when depicting the parameters necessary for action chunking.

We observe in \autoref{fig:tree_start} that the tree starts with many redundant nodes, e.g., $x_4\geq 0.41$ and $x_4\geq-0.99$, which leads to multiple paths potentially covering the same behaviors. With the pruning, we see in \autoref{fig:tree_end} that each node has unique comparisons and does not waste logic decisions. Furthermore, the specific features chosen in the logic decision make sense with respect to the LL-H domain: $x_5$ is the angular velocity, $x_0$ is the $x$-position, and $x_2$ is the $x$-velocity; these are the most significant features in achieving success in LL-H.

\section{Information-Theoretic Tree Restructuring}
\label{sec:restructure}
Here we write out the full algorithm design behind the Information-Theoretic Tree Restructuring algorithm (ITTR).

\begin{center}
\begin{minipage}{0.6\linewidth}
\begin{algorithm}[H]
\caption{Information-Theoretic Tree Restructuring}
\label{alg:deep-prune}
\begin{algorithmic}
\STATE {\bfseries Input:} Tree $T$, $D=\{(\mathrm{s}_t,\mathrm{a}_t,\hat A_t)\}_{t=0}^{N}, \epsilon$
\IF{$n$ steps have passed since last deepen / prune, \\ currently more than $m$ global steps have passed, \\
and at least $E$ policy complexity estimates were collected}
\STATE $T^* \leftarrow T$
\IF{$\frac{1}{N}\sum \nabla I^\pi(\textrm{s};\textrm{a}) < \epsilon$ (eq. \ref{alg:deep-condition}) and rewards are low}
    \STATE // deepen
    \STATE $l^* = \text{argmax}_l \mathbb{E}\left[p(l) * 1/\text{Var}\left[(\mathrm{a}-\mu_l)\hat A\right]\right]$ (eq. \ref{alg:deep-leaf})
    \STATE $w_\text{new}\sim \mathcal{N}(\mu_w,\Sigma_w)$
    \STATE $b_\text{new}\sim \mathcal{N}(\mu_b,\Sigma_b)$
    \STATE $\alpha_\text{new}\sim \mathcal{N}(\mu_\alpha,\Sigma_\alpha)$
    \STATE Remove $l^*$ from $T^*$ 
    \STATE Create a new node with $w,b,\alpha$ and add to $T^*$
    \STATE Make copies of $l^*$: $l_1$ and $l_2$, and add to $T^*$
\ELSIF{$\exists v_i<k$ and the tree has more than two leaves}
    \STATE // prune
    \STATE $l^*=\text{argmin} \{v_1,\ldots,v_l\}$
    \STATE Remove $l^*$ from $T^*$ 
    \STATE Remove the parent node of $l^*$ from $T^*$
    \STATE Move $l^*$'s sibling leaf up in $T^*$
\ENDIF
\STATE $T \leftarrow T^*$
\ENDIF
\end{algorithmic}
\end{algorithm}
\end{minipage}
\end{center}

Note that we consider rewards to be low when the mean return in the current rollout buffer is below $0$ ($-0.3$ for LK) and, specifically for LL-H, when fewer than half of the returns are positive. Returns are computed from the normalized rewards. For LL-H, we check the pruning condition before the deepening condition to incentivize pruning more often. Lastly, for warm-started trees in LL-H, we disable deepening, since the distilled tree already provides the structure, and check for pruning every $n=1e6$ steps with $k=4e6$.

\section{Further explanation: Number of parameters after deepening / pruning}
\label{sec:further-restructure}

% Emphasize that the DDTs truly have significantly less parameters to justify our method.
\autoref{tab:num_params} depicts the number of parameters used in the action chunked DDTs that are only relevant to the tree flow logic, i.e., no leaf parameters. We observe that the number of parameters drastically decreases, occurring because the majority of parameters are situated in the leaves used for action chunking. Subsequently, parameters relevant to the logic flow within DDT methods only make up to, at most, a tenth of the number of parameters in the MLP methods, enabling significantly more interpretability and easier debugging.

% Obs / action space of each domain:
% LK: obs space (12), action space (1)
% IP: obs space (4), action space (1)
% LL: obs space (8), action space (2)
% SP: obs space (60), action space (2)
\begin{table}[h]
    \centering
   \caption{Number of parameters used only for the logic flow of DDTs across the four domains and three runs (mean$\pm$s.e.).}
    \begin{tabular}{c c c c c}
         \hline
         Algorithm & LK & IP & LL & LL-H \\
         \hline
         \textsc{DDT-Ensemble NoLeaves} & $18\pm0.0$ & $10\pm0.0$ & $14\pm0.0$ & $20.0\pm4.9$\\
         \textsc{DDT-Pred NoLeaves} & $25.3\pm6.0$ & $10\pm0.0$ & $14\pm0.0$ & $26.0\pm4.9$\\
         \textsc{Warm-Ensemble NoLeaves} & $18\pm0.0$ & $10\pm0.0$ & $14\pm0.0$ & $26.0\pm4.9$\\
         \textsc{Warm-Pred NoLeaves} & $18\pm0.0$ & $10\pm0.0$ & $14\pm0.0$ & $33.3\pm9.4$\\
         \hline
    \end{tabular}
    \label{tab:num_params}
\end{table}

\section{Ablations}
\label{sec:ablations}
We perform ablations over the EMA coefficient parameter. Intuitively, a higher coefficient gives each action within an ensemble the same weight, promoting smoother actions but potentially causing worsened performance. We achieve this through running five sweeps using the values $\eta\in\{0.5, 0.625, 0.75, 0.875, 1.0\}$.

\begin{figure*}[h!]
    \centering
    % --- First Row ---
    \begin{subfigure}[b]{0.48\textwidth}
        \centering
        \includegraphics[width=\textwidth]{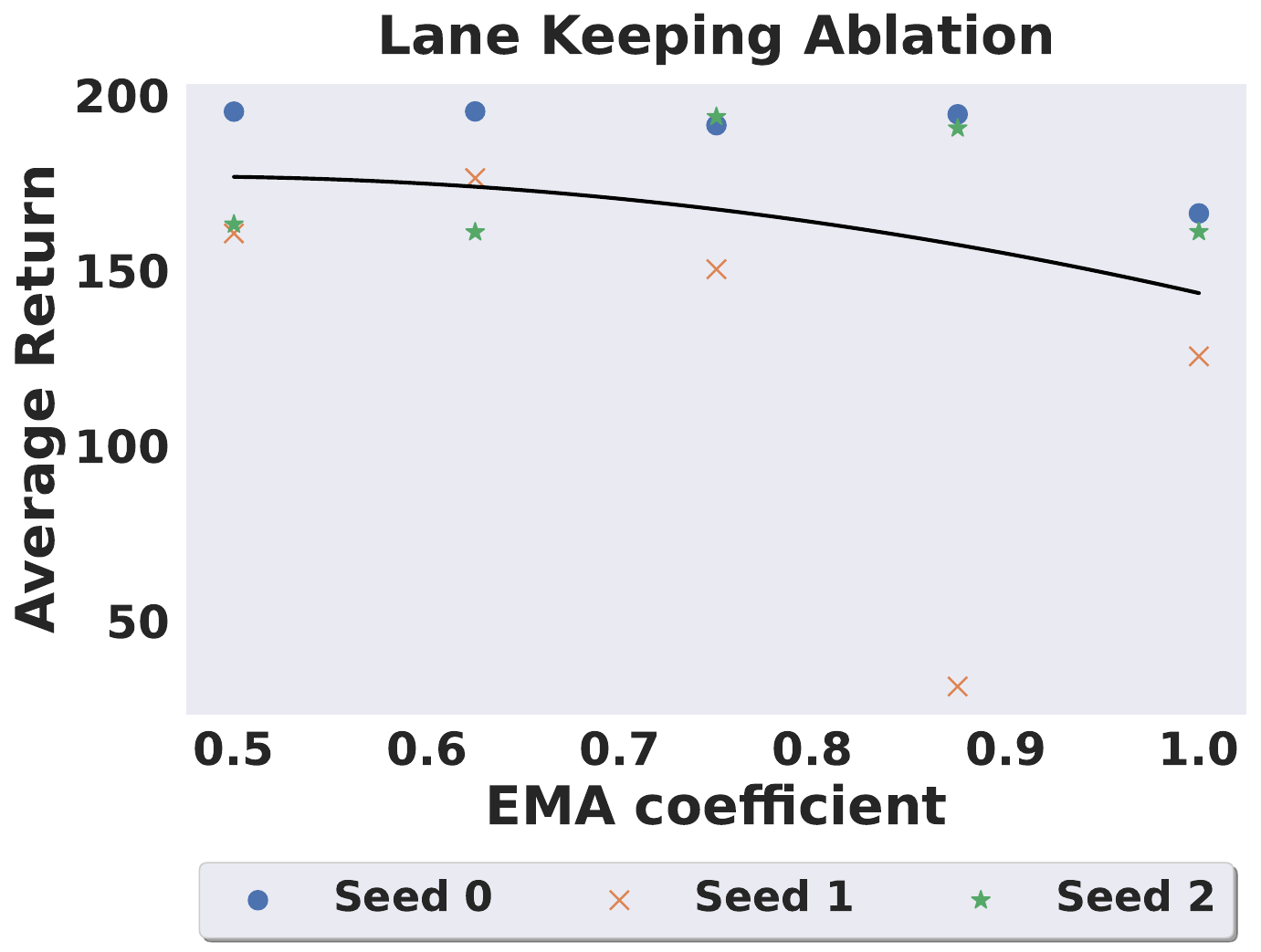}
        \caption{The trend in the performance against the EMA coefficient for the \textsc{DDT-Ensemble} algorithm in the \textsc{Lane Keeping} domain.}
        \label{fig:ablation-lk}
    \end{subfigure}
    \hfill
    \begin{subfigure}[b]{0.48\textwidth}
        \centering
        \includegraphics[width=\textwidth]{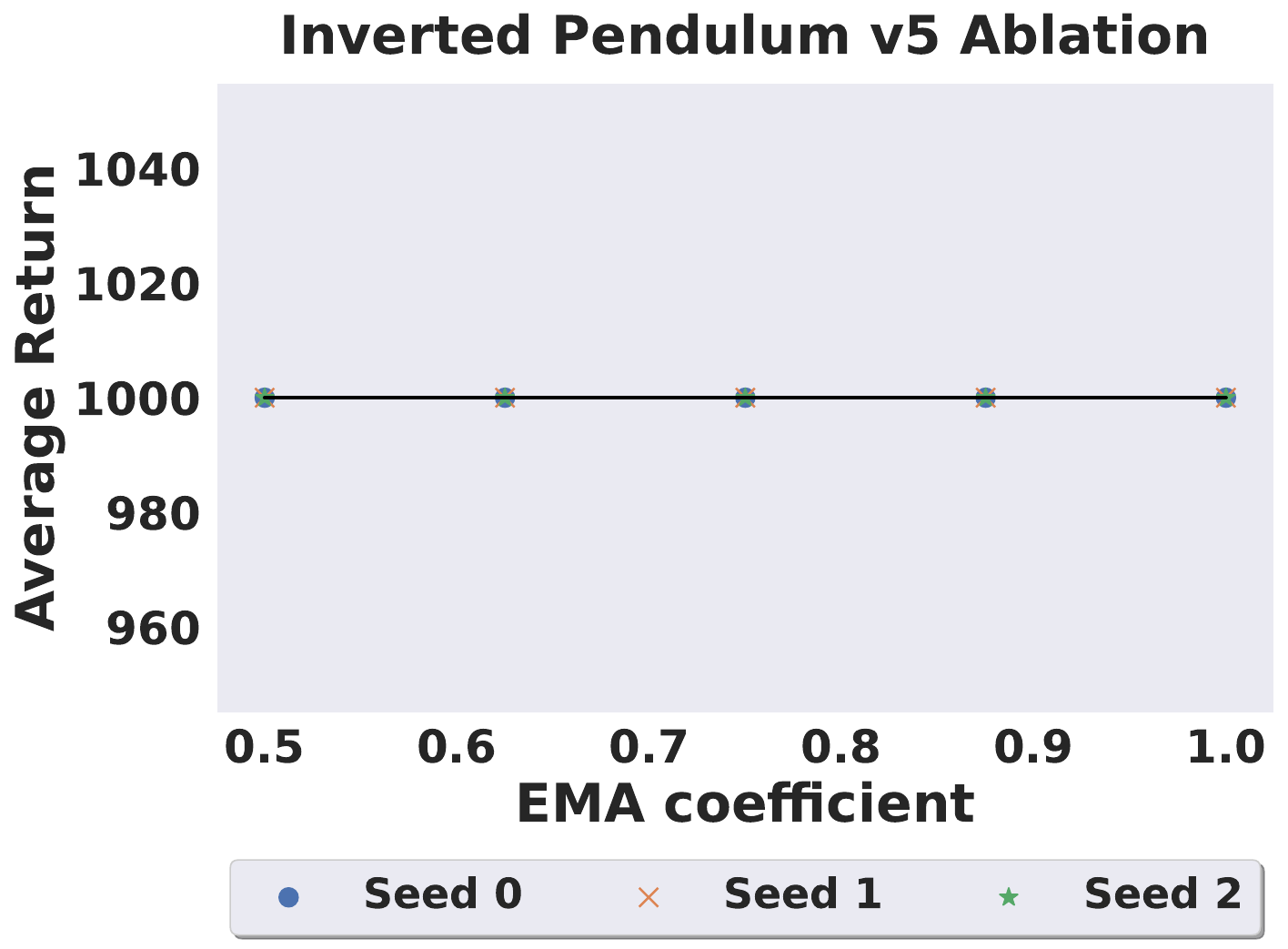}
        \caption{The trend in the performance against the EMA coefficient for the \textsc{DDT-Ensemble} algorithm in the \textsc{Inverted Pendulum} domain.}
        \label{fig:ablation-ip}
    \end{subfigure}

    % --- Second Row ---
    \begin{subfigure}[b]{0.48\textwidth}
        \centering
        \includegraphics[width=\textwidth]{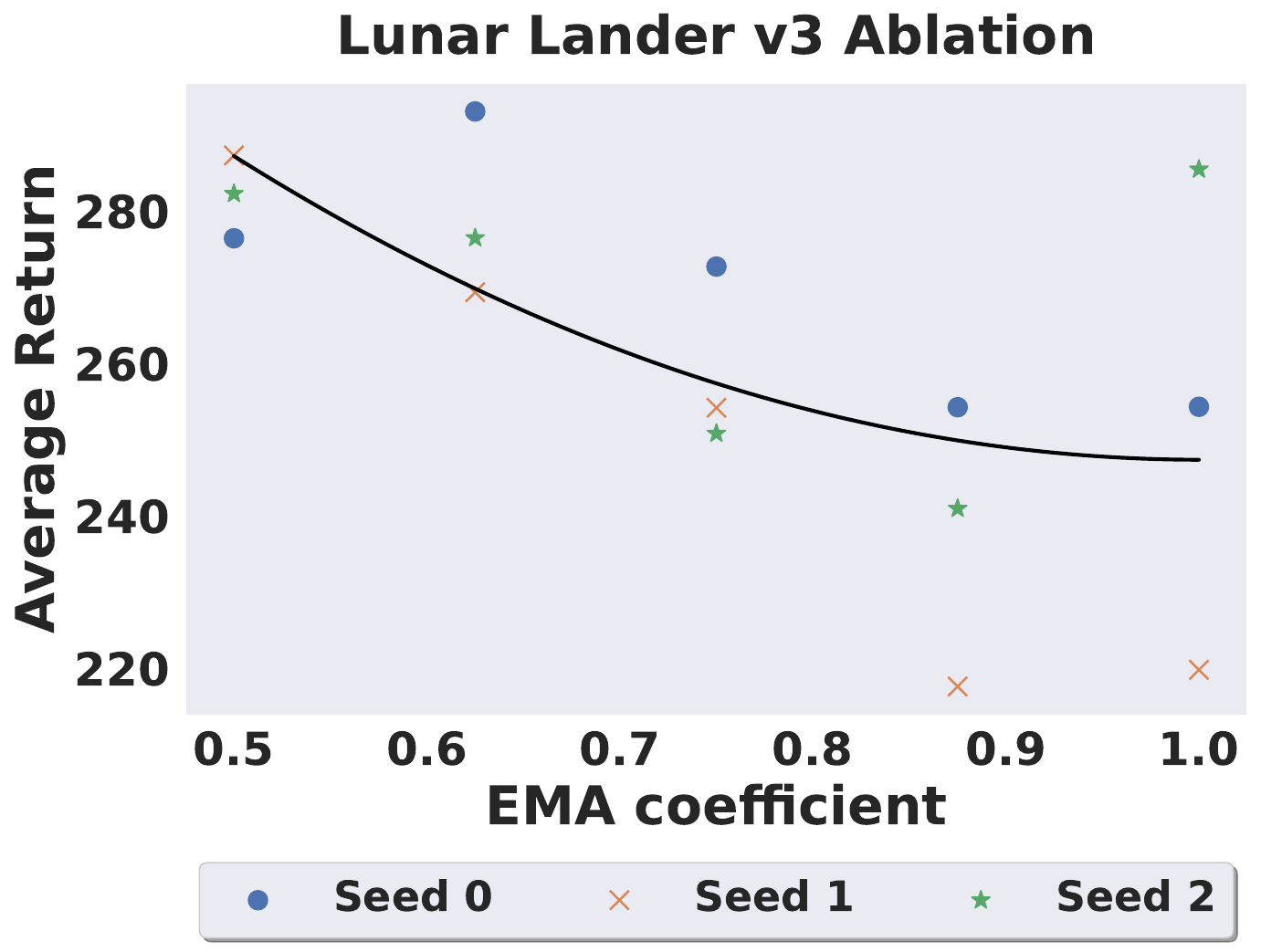}
        \caption{The trend in the performance against the EMA coefficient for the \textsc{DDT-Ensemble} algorithm in the \textsc{Lunar Lander v3} domain.}
        \label{fig:ablation-ll}
    \end{subfigure}
    \hfill
    \begin{subfigure}[b]{0.48\textwidth}
        \centering
        \includegraphics[width=\textwidth]{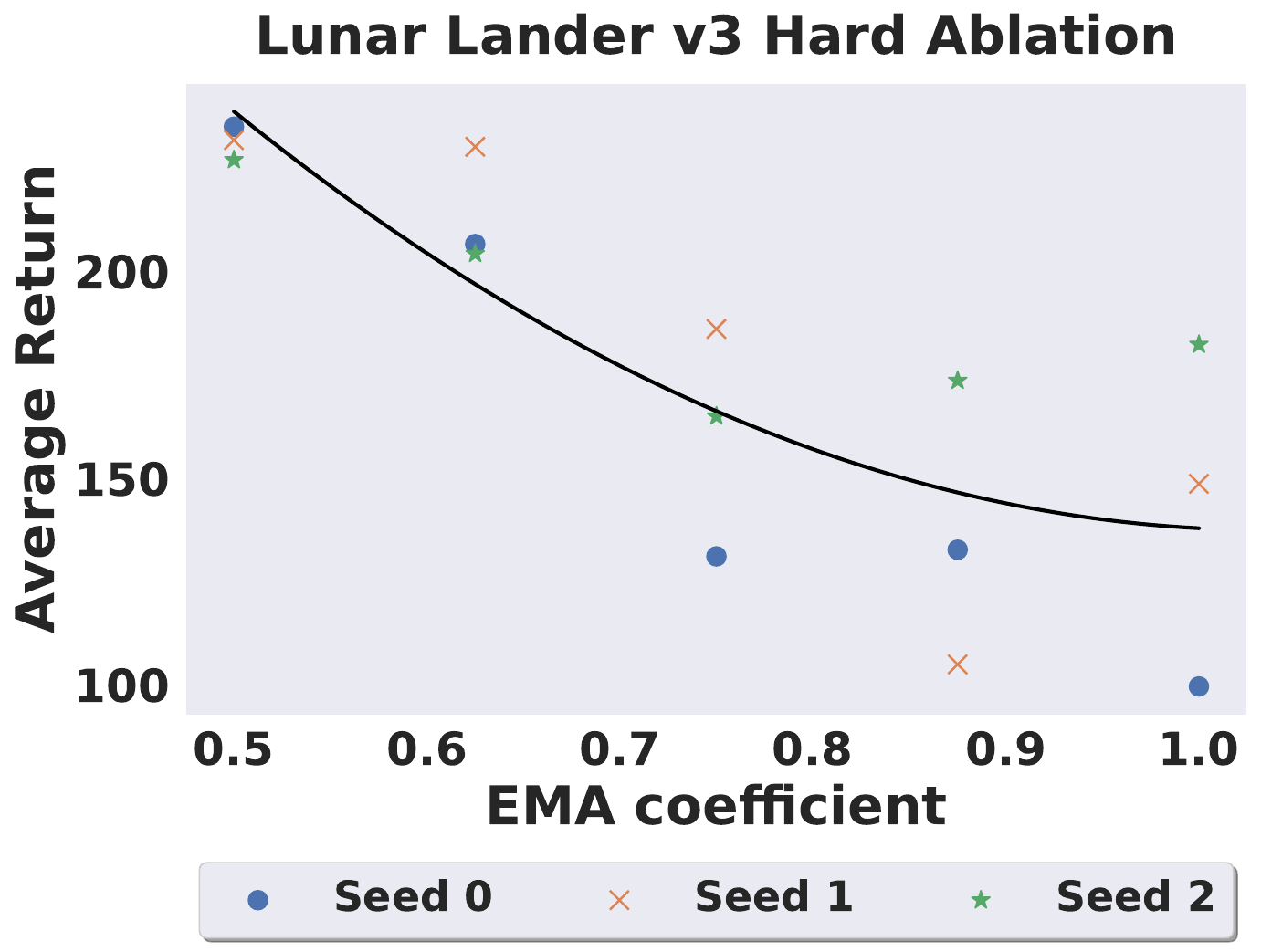}
        \caption{The trend in the performance against the EMA coefficient for the \textsc{DDT-Ensemble} algorithm in the \textsc{Lunar Lander Hard} domain.}
        \label{fig:ablation-ll-h}
    \end{subfigure}
    
    \caption{Figures depicting the trend in the performance against the EMA coefficient in each simulation environment. The black curve depicts the regression over the results across all seeds.}
    \label{fig:ablation-results}
\end{figure*}

\autoref{fig:ablation-results} depicts our results. In all domains except IP, performance decreases as the EMA coefficient increases, as expected according to our hypothesis. The decrease is especially apparent in the LL-H environment where the return drops from $231.2\pm1.9$ at $\eta=0.5$ to $143.6\pm19.6$ at $\eta=1.0$, likely due to the enforced smoothness of the actions performing poorly under LL-H's highly stochastic nature. Note that a lower EMA coefficient, in turn, gives the actions planned for later timesteps less weight, so while they contribute less to the aggregated action, they also are less temporally interpretable. The EMA coefficient therefore trades off performance against how strongly the agent commits to its shown plan. Interestingly, the unbiased variant ($\eta=1.0$, $\S4.1$) does not perform best, suggesting that the limiting factor is the enforced consistency rather than the bias of the estimator.

\section{Miscellaneous}
\label{sec:misc}

\subsection{Hyperparameters}
Here we report the hyperparameters used in all the experiments.

\begin{table}[h!]
  \caption{Shared hyperparameters used for all environments.}
  \label{table:hp-ddt-te}
  \begin{center}
        \begin{tabular}{lcccr}
          \toprule
          Parameter  & Value  \\
        \midrule
          \textsc{RPO} \\
          \qquad Actor learning rate & $5\textrm{e}{-4}$\\
          \qquad Critic learning rate & $5\textrm{e}{-4}$\\
          \qquad PPO clip rate & $0.2$\\
          \qquad RPO uniform bonus \cite{rahman2025robust} & $0.5$\\
          \midrule
          \textsc{MLP} \\
          \qquad Hidden layers & $[16,16]$\\
        \midrule
          \textsc{Temporal Ensemble} \\
          \qquad EMA coefficient $\eta$ & $0.75$\\
          
          \bottomrule
        \end{tabular}
  \end{center}
  \vskip -0.1in
\end{table}

\begin{table}[h!]
 \caption{Unique hyperparameters used for each environment for ITTR.}
  \label{table:hp-ddt-te}
  \begin{center}
   \begin{tabular}{lcccr}
       \toprule
        & \multicolumn{4}{c}{Domain} \\
        \cmidrule(lr){2-5}
         Parameter & LK & IP & LL & LL-H \\
         \midrule[1pt]
          \textsc{DDT} \\
         \qquad Starting $\#$ of leaves & $2$ & $2$ & $2$ & $4$ \\
         \qquad Maximum $\#$ of CART leaves & $2$ & $2$ & $2$ & $4$ \\
         \midrule[1pt]
         \textsc{Information-Theoretic Tree Restructuring} \\
         \qquad Minimum steps since last restructure $n$ & $200$ & $600$ & $1000$ & $1000$ \\
         \qquad Minimum global steps passed $m$ & $2000$ & $6000$ & $10000$ & $10000$ \\
         \qquad Minimum average policy complexity gradient $\epsilon$ & $5\mathrm{e}{-2}$ & $2\mathrm{e}{-1}$ & $5\mathrm{e}{-2}$ & $5\mathrm{e}{-3}$\\
          \qquad Number of visitations $k$ & $100000$ & $100000$ & $100000$ & $100000$\\
          \qquad Minimum number of policy complexity estimates $E$ & 2 & 3 & 4 & 4 \\
      \midrule[1pt]
          \textsc{Temporal Prediction} \\
          \qquad Scaling coefficient $\lambda$ & $1.0$ & $1.0$ & $1.0$ & $0.005$\\
         
         \bottomrule[1pt]
    \end{tabular}      
  \end{center}
  \vskip -0.1in
\end{table}

% \newpage
% \subsection{Choosing the Temporal Prediction Scaling Coefficient}
% We perform hyperparameter tuning for the scaling coefficient and find no noticeable difference in its weight. We thus initialize $\lambda=1.0$ for simplicity.

%%%%%%%%%%%%%%%%%%%%%%%%%%%%%%%%%%%%%%%%%%%%%%%%%%%%%%%%%%%%%%%%%%%%%%%%%%%%%%%
%%%%%%%%%%%%%%%%%%%%%%%%%%%%%%%%%%%%%%%%%%%%%%%%%%%%%%%%%%%%%%%%%%%%%%%%%%%%%%%

\end{document}